\RequirePackage{fix-cm}
\documentclass[twocolumn,natbib]{svjour3}          
\smartqed  

\usepackage{graphicx}
\usepackage{comment}
\usepackage{xspace}
\usepackage[caption=false]{subfig}
\newcommand{\ie}{\textit{i}.\textit{e}.\@\xspace}
\newcommand{\eg}{\textit{e}.\textit{g}.\@\xspace}
\newcommand{\cf}{\textit{c}.\textit{f}.\@\xspace}

\def\datasetname{SYNTHIA-SF}
\def\datasetnamefull{SYNTHIA-San Francisco}

\makeatletter \let\cl@chapter\relax \makeatother

\usepackage{url}

\usepackage{comment}
\newcommand\todo[1]{\textcolor{black}{#1}}
\newcommand\rev[1]{\textcolor{black}{#1}}
\newcommand\revs[1]{\textcolor{black}{#1}}
\usepackage{bm}
\usepackage{amssymb}
\usepackage{amsmath}
\usepackage{multirow}

\usepackage{adjustbox}
\usepackage{booktabs}
\usepackage[svgnames,table]{xcolor}

\newcommand{\punctspace}{\ }

\newcommand{\func}[2]{#1\mathopen{}\left(#2\mathclose{}\right)}

\newcommand{\pmf}[2][]{\func{P_{#1}}{#2}}

\newcommand{\upmf}[2][]{\func{\tilde{P}_{#1}}{#2}}
\newcommand{\given}{\mid}

\newcommand{\energyf}[2][]{\func{E_{#1}}{#2}}

\makeatletter

\newcommand{\minimize}{\@ifstar{\@minimizes}{\@minimize}}
\newcommand{\@minimizes}[1]{\ensuremath{ \operatorname{minimize } } }
\newcommand{\@minimize }[1]{\ensuremath{&\operatorname{minimize} &&}}
\newcommand{\maximize}{\@ifstar{\@maximizes}{\@maximize}}
\newcommand{\@maximizes}[1]{\ensuremath{ \operatorname{maximize } } }
\newcommand{\@maximize }[1]{\ensuremath{&\operatorname{maximize} &&}}
\newcommand{\minimizex}{\@ifstar{\@minimizexs}{\@minimizex}}
\newcommand{\@minimizexs}[1]{\ensuremath{ \underset{#1}{\operatorname{minimize}}\ }}
\newcommand{\@minimizex }[1]{\ensuremath{&\underset{#1}{\operatorname{minimize}}&&}}
\newcommand{\maximizex}{\@ifstar{\@maximizexs}{\@maximizex}}
\newcommand{\@maximizexs}[1]{\ensuremath{ \underset{#1}{\operatorname{maximize}}\ }}
\newcommand{\@maximizex }[1]{\ensuremath{&\underset{#1}{\operatorname{maximize}}&&}}
\newcommand{\argmin}{\@ifstar{\@argmins}{\@argmin}}
\newcommand{\@argmins}[1]{\ensuremath{ \operatorname{argmin } } }
\newcommand{\@argmin }[1]{\ensuremath{&\operatorname{argmin} &&}}
\newcommand{\argmax}{\@ifstar{\@argmaxs}{\@argmax}}
\newcommand{\@argmaxs}[1]{\ensuremath{ \operatorname{argmax } } }
\newcommand{\@argmax }[1]{\ensuremath{&\operatorname{argmax} &&}}
\newcommand{\argminx}{\@ifstar{\@argminxs}{\@argminx}}
\newcommand{\@argminxs}[1]{\ensuremath{ \underset{#1}{\operatorname{argmin}}\ }}
\newcommand{\@argminx }[1]{\ensuremath{&\underset{#1}{\operatorname{argmin}}&&}}
\newcommand{\argmaxx}{\@ifstar{\@argmaxxs}{\@argmaxx}}
\newcommand{\@argmaxxs}[1]{\ensuremath{ \underset{#1}{\operatorname{argmax}}\ }}
\newcommand{\@argmaxx }[1]{\ensuremath{&\underset{#1}{\operatorname{argmax}}&&}}

\makeatother
\renewcommand{\vec}[1]{\ensuremath{\bm{\MakeLowercase{#1}}}}

\newcommand{\imgsymb}{I}

\newcommand{\col}{u}
\newcommand{\row}{v}

\makeatletter

\newcommand{\stixelsymb}{s}
\newcommand{\stixelidx}{i}

\newcommand{\stixel}{\@ifstar{\@stixels}{\@stixel}}
\newcommand{\@stixels}[1]{ \bm{\MakeLowercase{\stixelsymb}}_{#1}  }
\newcommand{\@stixel }[1]{ \bm{\MakeUppercase{\stixelsymb}}_{#1}  }

\newcommand{\stixeld}{\@ifstar{\@stixelds}{\@stixeld}}
\newcommand{\@stixelds}{ \stixel*{\stixelidx}  }
\newcommand{\@stixeld }{ \stixel {\stixelidx}  }

\newcommand{\stixelu}{\@ifstar{\@stixelus}{\@stixelu}}
\newcommand{\@stixelus}[1]{ \stixel*{\col #1}   }
\newcommand{\@stixelu }[1]{ \stixel {\col #1}   }

\newcommand{\stixelud}{\@ifstar{\@stixeluds}{\@stixelud}}
\newcommand{\@stixeluds}{ \stixelu*{\stixelidx}  }
\newcommand{\@stixelud }{ \stixelu {\stixelidx}  }

\newcommand{\stixelcol}{\@ifstar{\@stixelcols}{\@stixelcol}}
\newcommand{\@stixelcols}{ \bm{\MakeLowercase{\stixelsymb}}_{:}  }
\newcommand{\@stixelcol }{ \bm{\MakeUppercase{\stixelsymb}}_{:}  }

\newcommand{\stixelcolu}{\@ifstar{\@stixelcolus}{\@stixelcolu}}
\newcommand{\@stixelcolus}{ \bm{\MakeLowercase{\stixelsymb}}_{\col :}  }
\newcommand{\@stixelcolu }{ \bm{\MakeUppercase{\stixelsymb}}_{\col :}  }

\newcommand{\numstixelssymb}{n}

\newcommand{\numstixels}{\@ifstar{\@numstixelss}{\@numstixels}}
\newcommand{\@numstixelss}{ \MakeLowercase{\numstixelssymb}  }
\newcommand{\@numstixels }{ \MakeUppercase{\numstixelssymb}  }

\newcommand{\numstixelsu}{\@ifstar{\@numstixelsus}{\@numstixelsu}}
\newcommand{\@numstixelsus}{ {\numstixels*}_{\col}  }
\newcommand{\@numstixelsu }{ {\numstixels }_{\col}  }

\newcommand{\rowsymb}{v}
\newcommand{\rowtopsymb}{t}
\newcommand{\rowbotsymb}{b}

\newcommand{\rowtop}{\@ifstar{\@rowtops}{\@rowtop}}
\newcommand{\@rowtops}[1]{ \MakeLowercase{\rowsymb}_{#1}^{\text{\rowtopsymb}}  }
\newcommand{\@rowtop }[1]{ \MakeUppercase{\rowsymb}_{#1}^{\text{\rowtopsymb}}  }

\newcommand{\rowtopd}{\@ifstar{\@rowtopds}{\@rowtopd}}
\newcommand{\@rowtopds}{ \rowtop*{\stixelidx}  }
\newcommand{\@rowtopd }{ \rowtop {\stixelidx}  }

\newcommand{\rowtopu}{\@ifstar{\@rowtopus}{\@rowtopu}}
\newcommand{\@rowtopus}[1]{ \rowtop*{\col #1}  }
\newcommand{\@rowtopu }[1]{ \rowtop {\col #1}  }

\newcommand{\rowtopud}{\@ifstar{\@rowtopuds}{\@rowtopud}}
\newcommand{\@rowtopuds}{ \rowtopu*{\stixelidx}  }
\newcommand{\@rowtopud }{ \rowtopu {\stixelidx}  }

\newcommand{\rowbot}{\@ifstar{\@rowbots}{\@rowbot}}
\newcommand{\@rowbots}[1]{ \MakeLowercase{\rowsymb}_{#1}^{\text{\rowbotsymb}}  }
\newcommand{\@rowbot }[1]{ \MakeUppercase{\rowsymb}_{#1}^{\text{\rowbotsymb}}  }

\newcommand{\rowbotd}{\@ifstar{\@rowbotds}{\@rowbotd}}
\newcommand{\@rowbotds}{ \rowbot*{\stixelidx}  }
\newcommand{\@rowbotd }{ \rowbot {\stixelidx}  }

\newcommand{\rowd}{\@ifstar{\@rowds}{\@rowd}}
\newcommand{\@rowds}{ \MakeLowercase{\rowsymb}_{\stixelidx}  }
\newcommand{\@rowd }{ \MakeUppercase{\rowsymb}_{\stixelidx}  }

\newcommand{\rowbotu}{\@ifstar{\@rowbotus}{\@rowbotu}}
\newcommand{\@rowbotus}[1]{ \rowbot*{\col #1}  }
\newcommand{\@rowbotu }[1]{ \rowbot {\col #1}  }

\newcommand{\rowbotud}{\@ifstar{\@rowbotuds}{\@rowbotud}}
\newcommand{\@rowbotuds}{ \rowbotu*{\stixelidx}  }
\newcommand{\@rowbotud }{ \rowbotu {\stixelidx}  }

\newcommand{\classsymb}{c}

\newcommand{\groundsymb}{g}
\newcommand{\ground}{\mathcal{\MakeUppercase{\groundsymb}}}
\newcommand{\objectsymb}{o}
\newcommand{\object}{\mathcal{\MakeUppercase{\objectsymb}}}
\newcommand{\skysymb}{s}
\newcommand{\sky}{\mathcal{\MakeUppercase{\skysymb}}}

\newcommand{\class}{\@ifstar{\@classs}{\@class}}
\newcommand{\@classs}[1]{ \MakeLowercase{\classsymb}_{#1}  }
\newcommand{\@class }[1]{ \MakeUppercase{\classsymb}_{#1}  }

\newcommand{\classd}{\@ifstar{\@classds}{\@classd}}
\newcommand{\@classds}{ \class*{\stixelidx}  }
\newcommand{\@classd }{ \class {\stixelidx}  }

\newcommand{\classu}{\@ifstar{\@classus}{\@classu}}
\newcommand{\@classus}[1]{ \class*{\col #1}  }
\newcommand{\@classu }[1]{ \class {\col #1}  }

\newcommand{\classud}{\@ifstar{\@classuds}{\@classud}}
\newcommand{\@classuds}{ \classu*{\stixelidx}  }
\newcommand{\@classud }{ \classu {\stixelidx}  }

\newcommand{\dispattrsymb}{g}

\newcommand{\dispattr}{\@ifstar{\@dispattrs}{\@dispattr}}
\newcommand{\@dispattrs}[1]{ \MakeLowercase{\dispattrsymb}_{#1}  }
\newcommand{\@dispattr }[1]{ \MakeUppercase{\dispattrsymb}_{#1}  }

\newcommand{\dispattrd}{\@ifstar{\@dispattrds}{\@dispattrd}}
\newcommand{\@dispattrds}{ \dispattr*{\stixelidx}  }
\newcommand{\@dispattrd }{ \dispattr {\stixelidx}  }

\newcommand{\dispattru}{\@ifstar{\@dispattrus}{\@dispattru}}
\newcommand{\@dispattrus}[1]{ \dispattr*{\col #1}  }
\newcommand{\@dispattru }[1]{ \dispattr {\col #1}  }

\newcommand{\dispattrud}{\@ifstar{\@dispattruds}{\@dispattrud}}
\newcommand{\@dispattruds}{ \dispattru*{\stixelidx}  }
\newcommand{\@dispattrud }{ \dispattru {\stixelidx}  }

\newcommand{\imgattrsymb}{o}

\newcommand{\imgattr}{\@ifstar{\@imgattrs}{\@imgattr}}
\newcommand{\@imgattrs}[1]{ \MakeLowercase{\imgattrsymb}_{#1}  }
\newcommand{\@imgattr }[1]{ \MakeUppercase{\imgattrsymb}_{#1}  }

\newcommand{\imgattrd}{\@ifstar{\@imgattrds}{\@imgattrd}}
\newcommand{\@imgattrds}{ \imgattr*{\stixelidx}  }
\newcommand{\@imgattrd }{ \imgattr {\stixelidx}  }

\newcommand{\imgattru}{\@ifstar{\@imgattrus}{\@imgattru}}
\newcommand{\@imgattrus}[1]{ \imgattr*{\col #1}  }
\newcommand{\@imgattru }[1]{ \imgattr {\col #1}  }

\newcommand{\imgattrud}{\@ifstar{\@imgattruds}{\@imgattrud}}
\newcommand{\@imgattruds}{ \imgattru*{\stixelidx}  }
\newcommand{\@imgattrud }{ \imgattru {\stixelidx}  }

\newcommand{\dispmeassymb}{D}

\newcommand{\disp}{\@ifstar{\@disps}{\@disp}}
\newcommand{\@disps}[1]{ \MakeLowercase{\dispmeassymb}_{#1}  }
\newcommand{\@disp }[1]{ \MakeUppercase{\dispmeassymb}_{#1}  }

\newcommand{\dispd}{\@ifstar{\@dispds}{\@dispd}}
\newcommand{\@dispds}{ \disp*{\row}  }
\newcommand{\@dispd }{ \disp {\row}  }

\newcommand{\dispu}{\@ifstar{\@dispus}{\@dispu}}
\newcommand{\@dispus}[1]{ \disp*{\col #1}  }
\newcommand{\@dispu }[1]{ \disp {\col #1}  }

\newcommand{\dispud}{\@ifstar{\@dispuds}{\@dispud}}
\newcommand{\@dispuds}{ \dispu*{\row}  }
\newcommand{\@dispud }{ \dispu {\row}  }

\newcommand{\dispcol}{\@ifstar{\@dispcols}{\@dispcol}}
\newcommand{\@dispcols}{ \bm{\MakeLowercase{\dispmeassymb}}_{:}  }
\newcommand{\@dispcol }{ \bm{\MakeUppercase{\dispmeassymb}}_{:}  }

\newcommand{\dispcolu}{\@ifstar{\@dispcolus}{\@dispcolu}}
\newcommand{\@dispcolus}{ \bm{\MakeLowercase{\dispmeassymb}}_{\col :}  }
\newcommand{\@dispcolu }{ \bm{\MakeUppercase{\dispmeassymb}}_{\col :}  }

\newcommand{\confmeassymb}{C}
\newcommand{\conf}{\@ifstar{\@confs}{\@conf}}
\newcommand{\@confs}[1]{ \MakeLowercase{\confmeassymb}_{#1}  }
\newcommand{\@conf }[1]{ \MakeUppercase{\confmeassymb}_{#1}  }

\newcommand{\confd}{\@ifstar{\@confds}{\@confd}}
\newcommand{\@confds}{ \conf*{\row}  }
\newcommand{\@confd }{ \conf {\row}  }

\newcommand{\imgval}{\@ifstar{\@imgvals}{\@imgval}}
\newcommand{\@imgvals}[1]{ \MakeLowercase{\imgsymb}_{#1}  }
\newcommand{\@imgval }[1]{ \MakeUppercase{\imgsymb}_{#1}  }

\newcommand{\imgvald}{\@ifstar{\@imgvalds}{\@imgvald}}
\newcommand{\@imgvalds}{ \imgval*{\row}  }
\newcommand{\@imgvald }{ \imgval {\row}  }

\newcommand{\imgvalu}{\@ifstar{\@imgvalus}{\@imgvalu}}
\newcommand{\@imgvalus}[1]{ \imgval*{\col #1}  }
\newcommand{\@imgvalu }[1]{ \imgval {\col #1}  }

\newcommand{\imgvalud}{\@ifstar{\@imgvaluds}{\@imgvalud}}
\newcommand{\@imgvaluds}{ \imgvalu*{\row}  }
\newcommand{\@imgvalud }{ \imgvalu {\row}  }

\newcommand{\imgcol}{\@ifstar{\@imgcols}{\@imgcol}}
\newcommand{\@imgcols}{ \bm{\MakeLowercase{\imgsymb}}_{:}  }
\newcommand{\@imgcol }{ \bm{\MakeUppercase{\imgsymb}}_{:}  }

\newcommand{\imgcolu}{\@ifstar{\@imgcolus}{\@imgcolu}}
\newcommand{\@imgcolus}{ \bm{\MakeLowercase{\imgsymb}}_{\col :}  }
\newcommand{\@imgcolu }{ \bm{\MakeUppercase{\imgsymb}}_{\col :}  }

\newcommand{\pixclasssymb}{L}

\newcommand{\pixclassval}{\@ifstar{\@pixclassvals}{\@pixclassval}}
\newcommand{\@pixclassvals}[1]{ \MakeLowercase{\pixclasssymb}_{#1}  }
\newcommand{\@pixclassval }[1]{ \MakeUppercase{\pixclasssymb}_{#1}  }

\newcommand{\pixclassvalvec}{\@ifstar{\@pixclassvalvecs}{\@pixclassvalvec}}
\newcommand{\@pixclassvalvecs}[1]{ \vec{\MakeLowercase{\pixclasssymb}}_{#1}  }
\newcommand{\@pixclassvalvec }[1]{ \vec{\MakeUppercase{\pixclasssymb}}_{#1}  }

\newcommand{\pixclassvald}{\@ifstar{\@pixclassvalds}{\@pixclassvald}}
\newcommand{\@pixclassvalds}{ \pixclassval*{\row}  }
\newcommand{\@pixclassvald }{ \pixclassval {\row}  }

\newcommand{\pixclassvaldvec}{\@ifstar{\@pixclassvaldvecs}{\@pixclassvaldvec}}
\newcommand{\@pixclassvaldvecs}{ \pixclassvalvec*{\row}  }
\newcommand{\@pixclassvaldvec }{ \pixclassvalvec {\row}  }

\newcommand{\pixclasscol}{\@ifstar{\@pixclasscols}{\@pixclasscol}}
\newcommand{\@pixclasscols}{ \bm{\MakeLowercase{\pixclasssymb}}_{:}  }
\newcommand{\@pixclasscol }{ \bm{\MakeUppercase{\pixclasssymb}}_{:}  }

\newcommand{\pixclasscolu}{\@ifstar{\@pixclasscolus}{\@pixclasscolu}}
\newcommand{\@pixclasscolus}{ \bm{\MakeLowercase{\pixclasssymb}}_{\col :}  }
\newcommand{\@pixclasscolu }{ \bm{\MakeUppercase{\pixclasssymb}}_{\col :}  }

\newcommand{\cutpriorsymb}{C}

\newcommand{\cutpriorval}{\@ifstar{\@cutpriorvals}{\@cutpriorval}}
\newcommand{\@cutpriorvals}[1]{ \MakeLowercase{\cutpriorsymb}_{#1}  }
\newcommand{\@cutpriorval }[1]{ \MakeUppercase{\cutpriorsymb}_{#1}  }

\newcommand{\cutpriorvalvec}{\@ifstar{\@cutpriorvalvecs}{\@cutpriorvalvec}}
\newcommand{\@cutpriorvalvecs}[1]{ \vec{\MakeLowercase{\cutpriorsymb}}_{#1}  }
\newcommand{\@cutpriorvalvec }[1]{ \vec{\MakeUppercase{\cutpriorsymb}}_{#1}  }

\newcommand{\cutpriorvald}{\@ifstar{\@cutpriorvalds}{\@cutpriorvald}}
\newcommand{\@cutpriorvalds}{ \cutpriorval*{\row}  }
\newcommand{\@cutpriorvald }{ \cutpriorval {\row}  }

\newcommand{\cutpriorvaldvec}{\@ifstar{\@cutpriorvaldvecs}{\@cutpriorvaldvec}}
\newcommand{\@cutpriorvaldvecs}{ \cutpriorvalvec*{\row}  }
\newcommand{\@cutpriorvaldvec }{ \cutpriorvalvec {\row}  }

\newcommand{\cutpriorcol}{\@ifstar{\@cutpriorcols}{\@cutpriorcol}}
\newcommand{\@cutpriorcols}{ \bm{\MakeLowercase{\cutpriorsymb}}_{:}  }
\newcommand{\@cutpriorcol }{ \bm{\MakeUppercase{\cutpriorsymb}}_{:}  }

\newcommand{\cutpriorcolu}{\@ifstar{\@cutpriorcolus}{\@cutpriorcolu}}
\newcommand{\@cutpriorcolus}{ \bm{\MakeLowercase{\cutpriorsymb}}_{\col :}  }
\newcommand{\@cutpriorcolu }{ \bm{\MakeUppercase{\cutpriorsymb}}_{\col :}  }

\newcommand{\meassymb}{M}

\newcommand{\meas}{\@ifstar{\@meass}{\@meas}}
\newcommand{\@meass}[1]{ \bm{\MakeLowercase{\meassymb}}_{#1}  }
\newcommand{\@meas }[1]{ \bm{\MakeUppercase{\meassymb}}_{#1}  }

\newcommand{\meascol}{\@ifstar{\@meascols}{\@meascol}}
\newcommand{\@meascols}{ \bm{\MakeLowercase{\meassymb}}_{:}  }
\newcommand{\@meascol }{ \bm{\MakeUppercase{\meassymb}}_{:}  }

\newcommand{\meascolu}{\@ifstar{\@meascolus}{\@meascolu}}
\newcommand{\@meascolus}{ \bm{\MakeLowercase{\meassymb}}_{\col :}  }
\newcommand{\@meascolu }{ \bm{\MakeUppercase{\meassymb}}_{\col :}  }

\newcommand{\pixmeasd}{\@ifstar{\@pixmeasds}{\@pixmeasd}}
\newcommand{\@pixmeasds}{ \meas*{\row}  }
\newcommand{\@pixmeasd }{ \meas {\row}  }

\newcommand{\priorcoltop}{\@ifstar{\@priorcoltops}{\@priorcoltop}}
\newcommand{\@priorcoltops}{ \bm{\MakeLowercase{\rowtopsymb}}_{:}  }
\newcommand{\@priorcoltop }{ \bm{\MakeUppercase{\rowtopsymb}}_{:}  }

\newcommand{\priorcoltopu}{\@ifstar{\@priorcoltopus}{\@priorcoltopu}}
\newcommand{\@priorcoltopus}{ \bm{\MakeLowercase{\rowtopsymb}}_{\col :}  }
\newcommand{\@priorcoltopu }{ \bm{\MakeUppercase{\rowtopsymb}}_{\col :}  }

\newcommand{\priorcolbot}{\@ifstar{\@priorcolbots}{\@priorcolbot}}
\newcommand{\@priorcolbots}{ \bm{\MakeLowercase{\rowbotsymb}}_{:}  }
\newcommand{\@priorcolbot }{ \bm{\MakeUppercase{\rowbotsymb}}_{:}  }

\newcommand{\priorcolbotu}{\@ifstar{\@priorcolbotus}{\@priorcolbotu}}
\newcommand{\@priorcolbotus}{ \bm{\MakeLowercase{\rowbotsymb}}_{\col :}  }
\newcommand{\@priorcolbotu }{ \bm{\MakeUppercase{\rowbotsymb}}_{\col :}  }

\newcommand{\partitionfunction}{Z}

\newcommand{\proboutlierdisp}{p_{\text{out}}}

\newcommand{\eprior}[1]{\energyf[prior]{#1}}
\newcommand{\edata}[1]{\energyf[data]{#1}}
\newcommand{\edepthmodel}[1]{\energyf[plane]{#1}}

\newcommand{\depthmodelsymb}{\mu}
\newcommand{\depthmodel}{\func{\depthmodelsymb{}}{\stixeld*,\row}}

\newcommand{\planeasymb}{a}
\newcommand{\planebsymb}{b}

\newcommand{\planea}{\@ifstar{\@planeas}{\@planea}}
\newcommand{\@planeas}{ \MakeLowercase{\planeasymb} }
\newcommand{\@planea }{ \MakeUppercase{\planeasymb} }

\newcommand{\planeb}{\@ifstar{\@planebs}{\@planeb}}
\newcommand{\@planebs}{ \MakeLowercase{\planebsymb} }
\newcommand{\@planeb }{ \MakeUppercase{\planebsymb} }

\makeatother

\usepackage{scalerel}

\usepackage{tikz}
\usetikzlibrary{calc}
\usetikzlibrary{shapes}
\usetikzlibrary{fit}
\usetikzlibrary{chains}
\usetikzlibrary{decorations.pathreplacing}
\usetikzlibrary{patterns}
\usetikzlibrary{intersections}

\usepackage{pgfplots}
\pgfplotsset{compat=newest}
\pgfplotsset{plot coordinates/math parser=false}
\usetikzlibrary{pgfplots.groupplots}

\usepackage{tabularx,ragged2e}

\newcolumntype{L}{>{\raggedright\arraybackslash}X}

\definecolor{sidewalk}{rgb}{0.953125,0.13671875,0.90625}
\definecolor{road}{rgb}{0.5,0.25,0.5}
\definecolor{traffic light}{rgb}{0.9765625,0.6640625,0.1171875}
\definecolor{traffic sign}{rgb}{0.859375,0.859375,0.}
\definecolor{vegetation}{rgb}{0.41796875,0.5546875,0.13671875}
\definecolor{person}{rgb}{0.859375,0.078125,0.234375}
\definecolor{car}{rgb}{0.,0.,0.5546875}
\definecolor{fence}{rgb}{0.7421875,0.59765625,0.59765625}

\definecolor{terrain}{rgb}{0.59375,0.98046875,0.59375}
\definecolor{building}{rgb}{0.2734375,0.2734375,0.2734375}
\definecolor{wall}{rgb}{0.3984375,0.3984375,0.609375}
\definecolor{rider}{rgb}{0.99609375,0.,0.}
\definecolor{truck}{rgb}{0.,0.,0.2734375}
\definecolor{bus}{rgb}{0.,0.234375,0.390625}
\definecolor{train}{rgb}{0.,0.3125,0.390625}
\definecolor{motorcycle}{rgb}{0.,0.,0.8984375}
\definecolor{bicycle}{rgb}{0.46484375,0.04296875,0.125}
\definecolor{sky}{rgb}{0.2734375,0.5078125,0.703125}
\definecolor{pole}{rgb}{0.5078125,0.5078125,0.5078125}
\definecolor{road lines}{rgb}{0.615686274509804,0.9176470588235294,0.19607843137254902}

\newcommand\labelcolor[1]{\textcolor{white}{\cellcolor{#1}{\scriptsize #1}}}
\newcommand\labelcolorb[1]{\cellcolor{#1}{\scriptsize #1}}

\usepackage{cleveref}

\begin{document}

\title{Slanted Stixels\thanks{* Both authors contributed equally.}
}
\subtitle{A way to represent steep streets}


\author{Daniel Hernandez-Juarez$^{*1}$        \and
        Lukas Schneider$^{*2}$				\and
        Pau Cebrian$^1$						\and
        Antonio Espinosa$^1$				\and
        David Vazquez$^4$					\and
        Antonio M. Lopez$^{1,3}$			\and
        Uwe Franke$^2$						\and
        Marc Pollefeys$^5$					\and
        Juan C. Moure$^1$
}

\authorrunning{Daniel Hernandez-Juarez \etal} 

\institute{Daniel Hernandez-Juarez \at
              \email{dhernandez0@gmail.com}
           \and
              $^1$Universitat Aut{\`o}noma de Barcelona (UAB), Barcelona, Spain
           \and
              $^2$Daimler AG R\&D, B{\"o}blingen, Germany
           \and
              $^3$Computer Vision Center (CVC), Barcelona, Spain
           \and
              $^4$Element AI, Montreal, Canada
           \and
              $^5$ETH Z{\"u}rich, Z{\"u}rich, Switzerland
}

\date{Received: date / Accepted: date}

\maketitle

\begin{abstract}
This work presents and evaluates a novel compact scene representation based on Stixels that infers geometric and semantic information. Our approach overcomes the previous rather restrictive geometric assumptions for Stixels by introducing a novel depth model to account for non-flat roads and slanted objects. Both semantic and depth cues are used jointly to infer the scene representation in a sound global energy minimization formulation.

Furthermore, a novel approximation scheme is introduced in order to significantly reduce the computational complexity of the Stixel algorithm, and then achieve real-time computation capabilities.
The idea is to first perform an over-segmentation of the image, discarding the unlikely Stixel cuts, and apply the algorithm only on the remaining Stixel cuts. This work presents a novel over-segmentation strategy based on a Fully Convolutional Network (FCN), which outperforms an approach based on using local extrema of the disparity map.

We evaluate the proposed methods in terms of semantic and geometric accuracy as well as run-time on four publicly available benchmark datasets. Our approach maintains accuracy on flat road scene datasets while improving substantially on a novel non-flat road dataset.
\end{abstract}

\section{Introduction}
\label{sec:intro}


Autonomous vehicles, advanced driver assistance systems, robots and other intelligent devices need to understand their environment. \revs{For this purpose, both geometric (distance) and semantic (classification) sources of information are useful. We want to represent these inputs in a very compact model and compute them in real-time to serve as a building block of higher-level modules, such as localization and planning.}



\begin{figure*}[ht]
\centering
\includegraphics[width=1\linewidth]{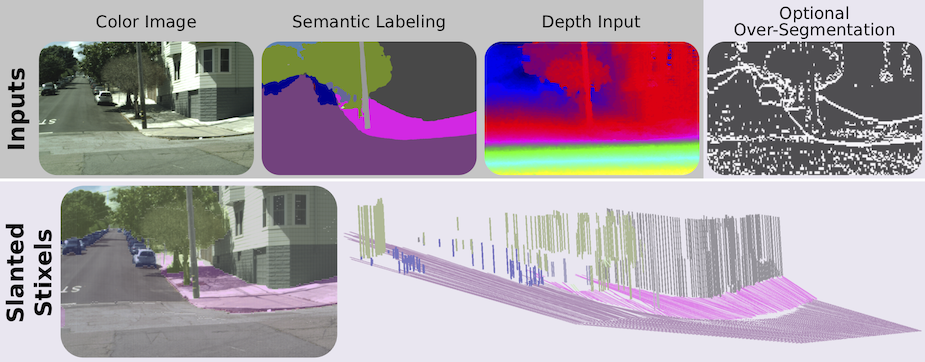}
\caption{The proposed approach: pixel-wise color, semantic and depth information serve as input to our Slanted Stixels model, which is a compact semantic representation of a 3D scene that accurately handles arbitrary scenarios such as San Francisco city. The optional over-segmentation in the top-right yields significant speed gains while nearly retaining the depth and semantic accuracy.}
\label{fig:overview}
\end{figure*}
This success has led to increased interest in the model from the intelligent vehicles community over the past years
The Stixel world has been successfully used for representing traffic scenes, as introduced in \cite{Pfeiffer2011}.
\rev{It has shown its potential particularly in the Bertha-Benz drive \citep{berthadrive}, where it has been successfully applied for visual scene understanding in autonomous driving. This success has led to increased interest in the model from the intelligent vehicles community over the past years} \citep{Schneider2016,Hernandez2017,Benenson2011,Cordts2014,Cordts2017,Oana2016,Levi2015,carrillo2016,HernandezSchneider2017}.

The Stixel world defines a compact medium-level representation of dense 3D disparity data obtained from stereo vision using rectangles, the so called \textit{Stixels}, as elements. \revs{Stixels are classified either as \textit{ground}-like planes, upright \textit{objects} or \textit{sky}, which are important geometric elements found in man-made environments.} This representation transforms millions of disparity pixels to hundreds or thousands of Stixels. At the same time, most scene structures, such as free space and obstacles, which are relevant for autonomous driving tasks, are adequately represented.

The idea behind the Stixel model is that planar surfaces are dominant in man-made environments and they can be modeled using this assumption. Scene structure found in urban environments can be modeled with certain constraints, \eg the sky is above the horizon line and objects usually lie on the ground. Generally, the geometric constraints of a scene are \revs{tied} to the vertical direction. Hence, the environment can be modeled as a column-wise segmentation of the image with a 3D stick-like shape, \ie a set of Stixels, \cf \cref{fig:overview}. The segmentation of the image is estimated by solving a column-wise energy minimization problem, taking depth and semantic cues as inputs as well as \textit{a priori} information that is used to regularize the solution \cf \cref{fig:overview}.

\rev{The Stixel model has been successfully used for automotive vision applications either to decrease parsing time, increase accuracy or both. We can find examples of works using the Stixel representation in different topics such as object recognition \citep{Benenson2012,LiFYXBPLG16}, building a grid map over time \citep{Muffert2014} and for autonomous driving \citep{berthadrive}.}
\rev{Specifically, for motion planning in the context of autonomous driving, the Stixel model has been used \cf \citep{berthadrive,ZieglerBDS14} to model the geometric constraints of a given scene.}

\begin{figure}
\subfloat[][Disparity representation of Stixels. The coloring encodes the distance from close (red) to far (green)]
{
\centering
\includegraphics[width=\linewidth]{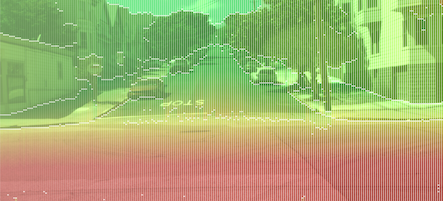}
\label{fig:stixel_img_disp}
}

\subfloat[][Semantic representation of Stixels. The coloring encodes the semantic class following \cite{Cordts2016}]
{
\centering
\includegraphics[width=\linewidth]{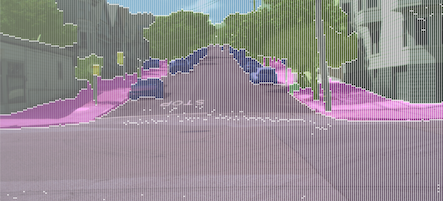}
\label{fig:stixel_img_sem}
}
\caption{Scene representation obtained by our method of a challenging street environment with a slanted road. Both geometric (top) and semantic (bottom) representations are shown.}
\label{fig:stixels_img}
\end{figure}

We propose a new depth model that is able to accurately represent arbitrary kinds of slanted objects and non-flat roads. The improved Stixel representation outperforms the original Stixel model in scenarios with non-flat roads, while keeping the same accuracy on flat road scenes. The induced extra computational complexity is reduced by incorporating an over-segmentation strategy that can be applied to any Stixel model proposed so far. An earlier version of our work \citep{HernandezSchneider2017} proposed a simple over-segmentation strategy that provided faster execution at the expense of decreasing the accuracy of the model.
This paper introduces a novel over-segmentation approach based on a Fully Convolutional Network (FCN) that outperforms the previous strategy, and achieves similar speedup results but retaining most of the accuracy of the original version.
An overview of our method is shown in \cref{fig:overview}.


\revs{Our main contributions are: (1) a novel depth model to accurately represent arbitrary kinds of slanted surfaces into the Stixel representation; (2) a novel over-segmentation prior designed to reduce the run-time of the method; (3) an effective over-segmentation strategy based on a shallow Fully Convolutional Network; (4) a new synthetic dataset with non-flat roads that includes pixel-level semantic and depth ground-truth, which is publicly available\footnotemark;
and (5) an in-depth evaluation in terms of run-time as well as semantic and depth accuracy carried out on this novel dataset and several real-world benchmarks.
Compared to the existing state-of-the-art approaches, our method substantially improves the depth accuracy in non-flat road scenarios.}

\footnotetext{http://synthia-dataset.net}

The remainder of this paper is structured as follows. \Cref{sec:related_work} reviews the state of the art. \Cref{sec:method} presents the new Stixel formulation.
We present two over-segmentation methods in \cref{sec:generate}.
\Cref{sec:experiments} explains the experiments we carried on and discusses their results. Finally, we state our conclusions in \cref{sec:conclusions}.

\section{Related work}
\label{sec:related_work}

Our proposed method introduces a novel Stixel-based scene representation that is able to account for non-flat roads, \cf \cref{fig:stixels_img}. We also devise an approximation to reduce the computational complexity of the underlying Dynamic Programming algorithm.

First, we will comment on works proposing different road scene models. Occupancy grid maps are models used to represent the surroundings of the vehicle~\citep{Dhiman2014,Muffert2014,Nuss2015,Thrun2002}. Typically, a grid in bird's eye perspective is defined and used to detect occupied grid cells and then, from this information, to extract the obstacles, drivable area, and unobservable areas from range data.
These grids and the Stixel world both represent the 2D image in terms of column-wise stripes allowing to capture the camera data in a polar fashion. Also, the Stixel data model is similar to the forward step usually found in occupancy grid maps \citep{Cordts2017}. However, the Stixel inference method in the image domain presents important differences compared to classical grid-based approaches.

Our work builds upon the proposal from \cite{Schneider2016}: they use semantic cues in addition to depth to extract a Stixel representation, which is able to provide a rich yet compact representation of the traffic scene. However, their model assumes a constant road slant and is therefore limited to flat road scenarios.
In contrast, our proposal overcomes this drawback by incorporating a novel plane model together with effective priors on the plane parameters.

Our proposal of using Stixels cuts is related to \cite{Cordts2014}: they use fast object detectors for different object classes, \eg Viola-Jones cascade detector \citep{viola2001rapid}, to produce top and bottom Stixel cuts that are used as prior information, which is then integrated into the Stixel algorithm. They prove that using object-level knowledge provides significant accuracy improvements.
Instead, we leverage semantic information as \rev{pixel}-level knowledge in our model for the same purpose of improving accuracy. Semantic segmentation identifies the objects and other elements of the image, \eg walls or sidewalks, providing pixel-level information, instead of boxes around the objects. Also, semantic segmentation requires a single predictor, while the method proposed by \cite{Cordts2014} needs a detector trained for each object class.
In contrast, we define a Stixel cut prior to generate an over-segmentation of the optimal Stixel cuts in order to speed up the execution of the algorithm.

There are some methods \citep{Benenson2011,Oana2016,Levi2015}, that represent simplified scene models with a single Stixel per column. The advantage of these approaches is that the computational complexity of the underlying algorithms is linear, but they cannot represent some complex scenarios found in the real world, \eg a pedestrian and a building in the same column.

Recent work by \cite{carrillo2016} uses edge-based disparity maps to compute Stixels. Their method is fast but they show that it gives inferior accuracy compared to the original Stixel model \citep{pfeiffer2013exploiting}.

\cite{Levi2015} firstly introduced the use of an FCN in Stixel-based methods. A single RGB image feeds the FCN to estimate
the bottom of the first non-road Stixel, \ie closest obstacle.
We use an FCN for a entirely different objective: to extract a Stixel cut over-segmentation that accelerates the execution of the algorithm. Moreover, the input of our FCN is a disparity map obtained from a stereo camera.

Finally, there are some works proposing fast implementations for Stixel computation. The FPGA implementation from \cite{Muffert2014} runs at 25 Hz with a Stixel width of 5 pixels, but the authors do not indicate the image resolution. \cite{Hernandez2017} present a GPU-accelerated implementation that runs at 26 Hz for a Stixel width of 5 pixels and image resolution of $1024\times440$ pixels, computed using a Semi-Global Matching (SGM) \citep{Hirschmuller2008} stereo algorithm.
We propose a novel approximation that accelerates the computation by reducing the algorithmic complexity. Accordingly, our proposal could benefit from the aforementioned FPGA- or GPU-accelerated implementations.

\section{Stixel Model}
\label{sec:method}

The Stixel world is a compressed representation of a 3D scene that preserves its relevant structure.
Since the structure in street environments is dominant in the vertical domain, the Stixel world leverages this idea to model a scene without taking into account the horizontal neighborhood.
This assumption leads to an efficient inference method and also allows the inference to be performed on all columns in parallel.

\begin{figure}[ht]
\centering
\includegraphics[width=1\linewidth]{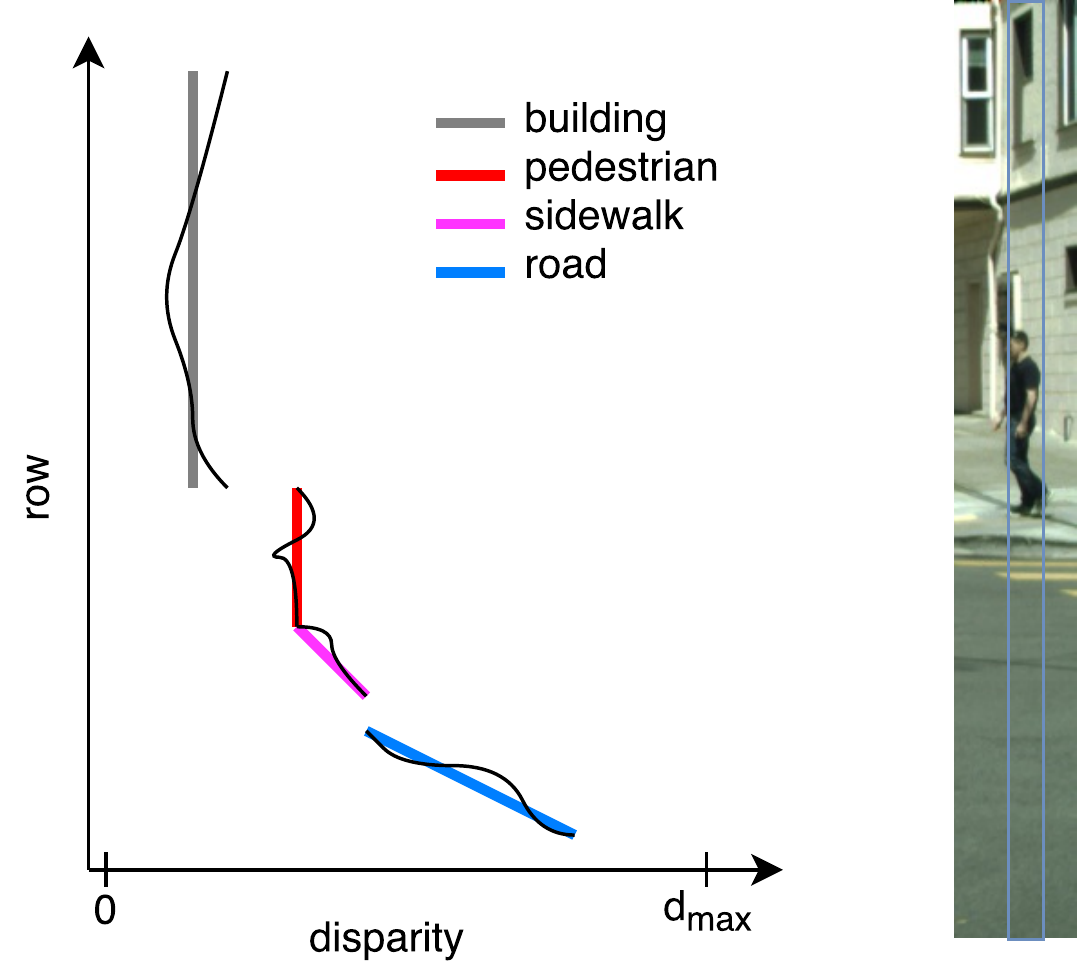}
\caption{Example of input disparity measurements (black lines) and output Stixels encoded with semantic colors (colored lines) for a typical scene column (right). Adapted from \cite{Cordts2017}.}
\label{fig:stixel_model_fig}
\end{figure}

The Stixel world is defined as a segmentation of image columns into stick-like super-pixels with class labels and a 3D planar depth model \cf \cref{fig:stixel_model_fig}.
We consider three structural classes: \textit{object}, \textit{ground} and \textit{sky}.
These classes have properties that are derived from an underlying 3D model: for \textit{object} Stixels the distance is roughly constant and usually lie on the ground, for \textit{sky} Stixels the distance is infinite and for \textit{ground} Stixels we favor planes with accordance to the expected ground.

The Stixel world has several properties that are useful for higher-level processing stages: (1) it is a medium-level scene representation that significantly reduces the quantity of elements, \eg from millions of pixels to hundreds of Stixels, while keeping an abstract representation of physical extent, depth and semantics; (2) the representation is based upon a street model; (3) the representation is not high-level because an object is represented by more than one Stixel horizontally and it can be split in more than one Stixel vertically too, \eg occlusions and slanted objects such as cars viewed from behind.

The joint Stixel segmentation and labeling problem is carried out via optimization of the column-wise posterior distribution $\pmf{\stixel{:} \given \meas{:} }$ defined over a Stixel segmentation $\stixel{:} $ given all measurements $\meas{:} $ from that particular image column. In the following, we drop the column indexes for ease of notation. We obtain Stixel width $>1$ as illustrated \eg in \cref{fig:overview} by down-sampling of the inputs, this width is fixed and is chosen to reduce the computational complexity during inference, however heavy down-sampling leads to degradation in accuracy \citep{Cordts2017}.

A Stixel column segmentation consists of an arbitrary number $\numstixels$ of Stixels $\stixel{i}$, each representing four random variables: the Stixel extent via bottom $\rowbotd$ and top $\rowtopd$ row, as well as its class $\classd$ and \rev{geometric depth model $\dispattrd$}. Thereby, the number of Stixels itself is a random variable that is optimized jointly during inference.
%
%
To this end, the posterior probability is defined by means of the unnormalized prior and likelihood distributions
\begin{equation}
  \pmf{\stixel{} \given \meas{}} =
        \frac{1}{\partitionfunction} \upmf{\meas{} \given \stixel{}} \upmf{\stixel{}} \punctspace
\label{eq:posterior}
\end{equation}
where $Z$ is the normalizing partition function. Transformed to log-likelihoods via
\begin{equation}
\pmf{\stixel{}=\stixel*{} \given \meas{}=\meas*{}} \allowbreak = -\log(e^{ - \energyf{\stixel*{},\meas*{}}})
\label{eq:logeq}
\end{equation}

where $\energyf{\cdot}$ is the energy function, $\edata{\cdot}$ is the \textbf{likelihood} term and $\eprior{\cdot}$ is the \textbf{prior} term.

\begin{equation}
\energyf{\stixel*{},\meas*{}} = \edata{\stixel*{},\meas*{}} + \eprior{\stixel*{}}
\label{eq:energyfunction}
\end{equation}

\subsection{Data term}

The \textbf{likelihood} term $\edata{\cdot}$ thereby rates how well the measurements $\pixmeasd*$ at pixel $\row$ fit to the overlapping Stixel $\stixeld*$
\begin{align}
\begin{split}
\edata{\stixel*{},\meas*{}} &=
\sum_{\stixelidx=1}^{\numstixels}
\energyf[stixel]{\stixeld*,\meas*{}}
\\
& = \sum_{\stixelidx=1}^{\numstixels}
\sum_{\row = \rowbotd*}^{\rowtopd*}
\energyf[pixel]{\stixeld*,\pixmeasd*} \punctspace .
\end{split}
\label{eq:data}
\end{align}
This pixel-wise energy is further split in a semantic and a depth term
\begin{equation}
\energyf[pixel]{\stixeld*,\pixmeasd*} =
\energyf[disp]{\stixeld*,\dispd*} + w_l \cdot \energyf[sem]{\stixeld*,\pixclassvald*} .
\end{equation}
The parameter $w_l$ controls the influence of the semantic data term. The input is provided by an FCN that delivers normalized semantic scores $l_v(c_i)$ with $\sum_{c_i}l_v(c_i) = 1$ for all classes $c_i$ at pixels $v$. The semantic energy favors semantic classes of the Stixel that fit to the observed pixel-level semantic input \citep{Schneider2016}. The semantic likelihood term is
\begin{equation}
\energyf[sem]{\stixeld*,\pixclassvald*} = -log(l_v(c_i)) \punctspace .
\end{equation}

\rev{The depth model is designed to represent the different characteristics of the different geometric classes,
\ie \textit{object}, \textit{ground} and \textit{sky} Stixels.
Furthermore, the model enforces multiple stacked Stixels in cases of objects with the same class but different depths.}

Our depth input is a dense disparity map, each pixel is assigned a disparity value or is masked as invalid \ie $d_v \in \{0 ... d_{max}, d_{invalid}\}$. The depth term is defined by means of a probabilistic and generative sensor model $\pmf[\row]{\cdot}$ that considers the accordance of the depth measurement $\dispd*$ at row $\row$ to the Stixel $\stixeld*$

\begin{equation}
\energyf[disp]{\stixeld*,\dispd*} =
- \func{\log}{\pmf[\row]{\dispd=\dispd* \given \stixeld=\stixeld*}} \punctspace .
\label{eq:data_depth}
\end{equation}

Invalid $d_{inv}$ disparity measurements have to be handled, therefore, a prior probability of a valid disparity value is defined as $p_{val}$

\begin{equation}
\pmf[\row]{\dispd \given \stixeld} =
\begin{cases}
p_{val} P_{v,val}(\dispd \given \stixeld) & \mbox{if } d_v \neq d_{inv}\\
(1-p_{val})	& \mbox{otherwise}
\end{cases}
\label{eq:dispmeasmodel}
\end{equation}

where $P_{v,val}(\dispd \given \stixeld)$ is the measurement model of valid disparities only. It is comprised of a constant outlier probability $\proboutlierdisp$ and a Gaussian sensor noise model for valid measurements with confidence $\confd*$

\begin{equation}
P_{v,val}(\dispd \given \stixeld) = \frac{\proboutlierdisp}{\partitionfunction_{U}}
  + \frac{1-\proboutlierdisp}{\func{\partitionfunction_{G}}{\stixeld*}} e^{ -
    \left(
        \frac{ \confd*  \left(\dispd*-\depthmodel\right)}{\func{\sigma}{\stixeld*}}
    \right)^2 }
\label{eq:dispmeasmodel2}
\end{equation}

that is centered at the expected disparity $\depthmodel$ given by the depth model of the Stixel, where $\partitionfunction_{U}$ and $\func{\partitionfunction_{G}}{\stixeld*}$ normalize the distributions. Similarly to \cite{pfeiffer2013exploiting}, we use the confidence of the depth estimates $\confd*$ to influence the shape of the distribution. The Gaussian models the typical disparity noise and the uniform distribution makes the model more robust to outliers, which is weighted by $\proboutlierdisp$. The standard deviation $\func{\sigma}{\stixeld*}$ models the noise of the stereo matching algorithm and depends on the class $c_i$.

\subsubsection{New depth model}
\rev{
The depth model defines the 3D outline of a Stixel using very few parameters per Stixel and reflects our assumptions on the surrounding scene.
Both, data term (\cf \cref{eq:dispmeasmodel2}) and priors (\cf \cref{sec:prior}) have a significant impact on the inferred depth model.
In previous formulations,
the three different geometric classes were designed using restrictive constant height (ground Stixels) and constant depth (object and sky Stixels), assumptions per Stixel, \eg for object Stixels: $ \depthmodel=constant $.
}

\rev{
This paper introduces a new plane depth model that relaxes the previous assumptions in favor of a more accurate depth representation.
The new model is formulated such that it nicely interacts with this well founded and experimentally validated depth sensor model.}
To this end, we formulate the depth model $\depthmodel$ using two random variables defining a plane in the disparity space
that evaluates to the disparity in row $\row$ via
%
\begin{equation}
\depthmodel = \planeb*_\stixelidx \cdot \row + \planea*_\stixelidx \punctspace .
\end{equation}

Note that we assume narrow Stixels and thus can neglect one plane parameter, \ie the roll.

\begin{figure*}[ht]
  \centering
      \begin{tabular}{@{}c@{}c@{}}
        \includegraphics[width=0.37\textwidth]{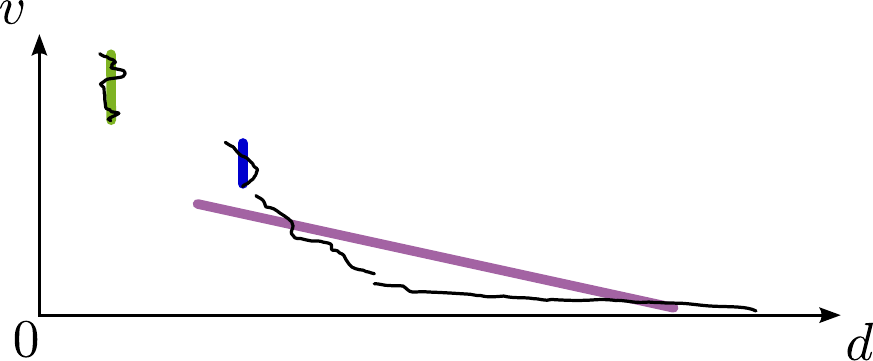} &
        \includegraphics[width=0.59\textwidth]{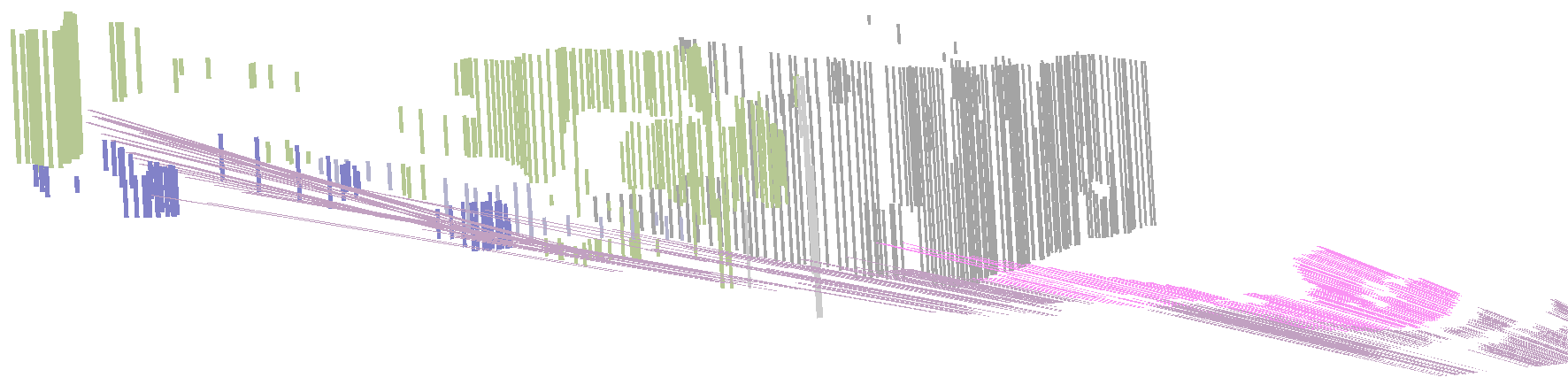} \\
        \includegraphics[width=0.37\textwidth]{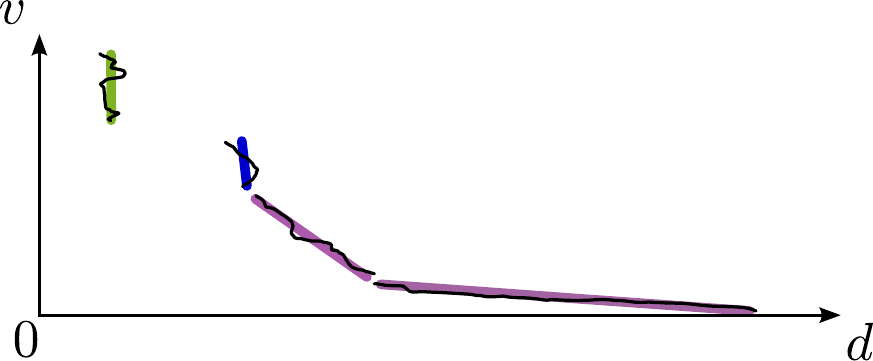} &
        \includegraphics[width=0.59\textwidth]{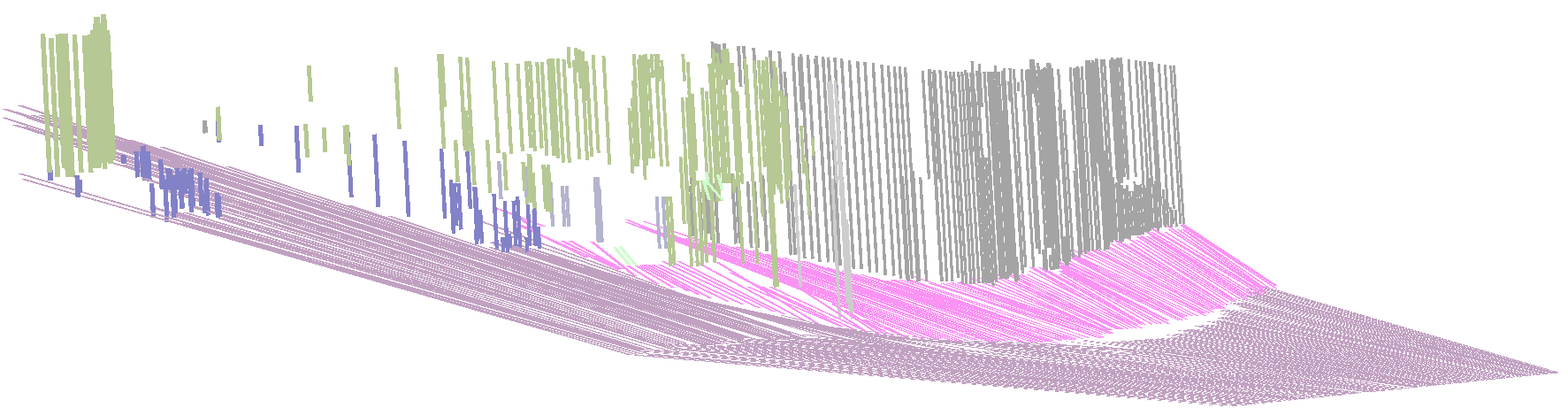}
      \end{tabular}
\caption{Comparison of original \citep{Schneider2016} (top) and our Slanted Stixels (bottom): due to the fixed slant in the original formulation, the road surface is not well represented as illustrated on the top-left figure. The novel model is capable of reconstructing the whole scene accurately.}
\label{fig:comparison_example}
\end{figure*}

This model is a generalization of the previous class-specific depth models used in previous works, allowing for a more flexible representation of the scene because of the extra free parameter \cf \cref{fig:comparison_example}. The way of modeling the different Stixel classes \ie \textit{object}, \textit{ground} and \textit{sky} is through priors, as explained in \cref{sec:plane_prior}. \rev{Also, to completely understand the details about the inference, we suggest to read \cref{sec:new_model}.}

\subsection{Prior term}
\label{sec:prior}

The \textbf{prior} captures knowledge about the segmentation independent from measurements, in this section we define the priors used for this model, they are based on \cite{Cordts2017}. The Markov property is used so that the prior reduces to pair-wise relations between subsequent Stixels. Accordingly, the prior is computed as
\begin{equation}
\eprior{\stixel*{}} =
\sum_{\stixelidx=2}^{\numstixels}
\energyf[pair]{\stixeld*,\stixel*{\stixelidx-1}} +
\energyf[first]{\stixel*{1}} \punctspace .
\label{eq:prior}
\end{equation}

\rev{In the next sections, where different priors are introduced, $\energyf[pair]{\stixeld*,\stixel*{\stixelidx-1}}$ is the summation of all these priors. However, $\energyf[first]{\stixel*{1}}$ does not include pairwise terms, \ie}

\rev{
\begin{align}
\begin{split}
\energyf[first]{\stixel*{1}} = & \energyf[mc]{\stixel*{1}} + \energyf[segfirst]{\stixel*{1}} + \energyf[seglast]{\stixel*{1}} \\
& + \energyf[top \geq bottom]{\stixeld*} + \edepthmodel{\stixel*{1}}
\end{split}
\end{align}
}

\subsubsection{Model complexity prior}
\label{subsubsec:model_complexity_prior}

A model complexity term favors solutions composed of fewer Stixels and thus invokes costs for each Stixel in the column segmentation $\stixel{}$:
\begin{equation}
\rev{\energyf[mc]{\stixeld*}}= C_{mc} \punctspace .
\end{equation}

There is a trade-off between compactness and accuracy. A high $C_{mc}$ parameter would lead to a very compact segmentation \ie few Stixels. However, a representation with few Stixels is more likely to have lower accuracy, \eg a solution comprised of one Stixel the size of the whole column would result in a huge disparity and semantic error.

\subsubsection{Segmentation priors}
The model has to enforce that all pixels are assigned to exactly one Stixel, \ie non-overlapping Stixels, Stixels extend over all the column and Stixels are connected. Therefore, the first priors are defined to comply with the following rules: The first Stixel must begin in row 1 and the last Stixel must end in row $h$, \ie

\begin{equation}
\rev{\energyf[segfirst]{\stixeld*}} =
\begin{cases}
\infty & \mbox{if } v_i^b \neq 1, \mbox{i} = 1 \\
0 & \mbox{otherwise}
\end{cases}
\label{eq:prior_first}
\end{equation}

\begin{equation}
\rev{\energyf[seglast]{\stixeld*}} =
\begin{cases}
\infty & \mbox{if } v_i^t \neq h, \mbox{i} = n \\
0 & \mbox{otherwise}
\end{cases} \punctspace .
\label{eq:prior_last}
\end{equation}

Furthermore, every Stixel must be connected to the next one and the Stixel top row must be greater than the bottom row, \ie

\begin{equation}
\rev{\energyf[connection]{\stixeld*,\stixel*{\stixelidx-1}}} =
\begin{cases}
0 & \mbox{if } v_i^b = v_{i-1}^t + 1 \\
\infty & \mbox{otherwise}
\end{cases}
\label{eq:connection}
\end{equation}

\begin{equation}
\rev{\energyf[top \geq bottom]{\stixeld*}} =
\begin{cases}
0 & \mbox{if } v_i^b \leq v_i^t \\
\infty & \mbox{otherwise}
\end{cases} \punctspace .
\label{eq:top_geq_bottom}
\end{equation}

\subsubsection{Structural priors}

The gravity prior penalizes a flying object \ie an \textit{object} Stixel not lying on top of the previous \textit{ground} Stixel,

\begin{equation}
\rev{\energyf[gravity]{\stixeld*,\stixel*{\stixelidx-1}}} =
\begin{cases}
\alpha_{gravity}^- + \beta_{gravity}^- \Delta_{\dispattrsymb} & \mbox{if } \Delta_{\dispattrsymb} < 0 \\
\alpha_{gravity}^+ + \beta_{gravity}^+ \Delta_{\dispattrsymb} & \mbox{if } \Delta_{\dispattrsymb} > 0 \\
0 & \mbox{otherwise}
\end{cases}
\label{eq:gravity}
\end{equation}

where $\Delta_{\dispattrsymb} = \depthmodelsymb_s(v_{i}^b, \dispattrsymb_{i}) - \depthmodelsymb_s(v_{i-1}^t,\dispattrsymb_{i-1})$ is the difference between the \textit{object} Stixel disparity  $\depthmodelsymb_s(v_{i}^b, \dispattrsymb_{i})$ at it's bottom pixel $v_i^b$ and the disparity of the \textit{ground} Stixel  $\depthmodelsymb_s(v_{i-1}^t, \dispattrsymb_{i-1})$ at the top row $v_{i}^t$. It only applies for $s_i$ being an object and $s_{i-1}$ being a ground Stixel.

The depth ordering prior penalizes a combination of two staggered \textit{object} Stixels when the upper of the two is closer (in distance to the car) than the lower one.

\begin{equation}
\rev{\energyf[ord]{\stixeld*,\stixel*{\stixelidx-1}}} =
\begin{cases}
\alpha_{ord} + \beta_{ord} (\dispattrsymb_i - \dispattrsymb_{i-1}) & \mbox{if } \dispattrsymb_i > \dispattrsymb_{i-1} \\
0 & \mbox{otherwise}
\end{cases} \punctspace .
\label{eq:ordering}
\end{equation}

A novel prior is introduced in this paper: the ground gap prior penalizes two consecutive \textit{ground} Stixels when the bottom disparity of the upper Stixel \ie disparity at row $v_{i}^b$ and the disparity of the lower Stixel at row $v_{i}^b$ do not match.

\begin{equation}
\rev{\energyf[gap]{\stixeld*,\stixel*{\stixelidx-1}}} =
\begin{cases}
\alpha_{gap}^- + \beta_{gap}^- \Delta_{gap} & \mbox{if } \Delta_{gap} < 0 \\
\alpha_{gap}^+ + \beta_{gap}^+ \Delta_{gap} & \mbox{if } \Delta_{gap} > 0 \\
0 & \mbox{otherwise}
\end{cases}
\label{eq:ground_gap}
\end{equation}

where $\Delta_{gap} = \dispattrsymb_s(v_i^b, \dispattrsymb_i) - \dispattrsymb_s(v_i^b, \dispattrsymb_{i-1})$. These structural priors do not enforce their assumptions. Instead, they penalize unusual combinations, \eg a flying object (gravity prior), traffic signs (ordering prior).

\subsubsection{Transition priors}

These priors define the knowledge regarding the transition between a pair of Stixels.

\begin{equation}
\rev{\energyf[transition]{\stixeld*,\stixel*{\stixelidx-1}}} =
\gamma_{c_i,c_{i-1}}
\label{eq:transition_prior}
\end{equation}

where $\gamma_{c_i,c_{i-1}}$ is the transition cost between previous Stixel class $c_{i-1}$ to current Stixel class $c_i$. This is defined via a two-dimensional transition matrix for all combinations of classes $\gamma_{c_i,c_{i-1}}$. Only first order relations are modeled in order to infer efficiently.

\subsubsection{Plane prior}
\label{sec:plane_prior}

In this paper, we propose a new additional prior term that uses the specific properties of the three geometric classes. We expect the two random variables $\planea , \planeb$ representing the plane parameters of a Stixel to be Gaussian distributed, \ie
\rev{
\begin{equation}
\rev{\edepthmodel{\stixeld*}} = \left(\frac{ \planea*_i - \mu_{\classd*}^{\planea*}  }
                                     {\sigma_{\classd*}^{\planea*}}           \right)^2
                          +
                          \left(\frac{ \planeb*_i - \mu_{\classd*}^{\planeb*}  }
                                     {\sigma_{\classd*}^{\planeb*}}           \right)^2
                          - \func{\log}{\partitionfunction} \punctspace .
\end{equation}
}

This prior favors planes in accordance to the expected 3D layout corresponding to the geometric class.  For instance, \textit{object} Stixels are expected to have an approximately constant disparity, \ie $\mu_{\textit{object}}^{\planeb*} = 0$. The expected road slant $\mu_{\textit{ground}}^{\planeb*}$ can be set using prior knowledge or a preceding road surface detection. For \textit{sky} Stixels we expect infinite distance \ie 0 disparity, therefore, we set $\mu_{\textit{sky}}^{\planea*} = \mu_{\textit{sky}}^{\planeb*} = 0$.

The standard deviations $\sigma_{\classd*}^{\planea*}$ and $\sigma_{\classd*}^{\planeb*}$ are used in order to enforce the assumptions for each Stixel class, \ie the more confident we are that \textit{object} Stixels have constant distance, the closer to 0 we would set $\sigma_{object}^{\planeb*}$. The same applies for \textit{ground} Stixels: if we know the road is not slanted, we can rely on the expected previous road model and set $\sigma_{ground}^{\planeb*} \rightarrow 0$. For \textit{sky} Stixels, it does not make sense to have a disparity different to 0. Therefore, we set $\sigma_{sky}^{\planea*} \rightarrow 0$ and $\sigma_{sky}^{\planeb*} \rightarrow 0$.

Note that the novel formulation is a strict generalization of the original method, since they are equivalent, \eg if the slant is fixed, \ie $\sigma_{object}^{\planeb*} \rightarrow 0, \mu_{object}^{\planeb*} = 0$.

\subsection{Inference}
\label{sec:new_model}
The sophisticated energy function defined in \cref{sec:method} is optimized via Dynamic Programming as in \cite{Pfeiffer2011}. However, we must also optimize jointly for the novel depth model. When optimizing for the plane parameters $\planea*_i,\planeb*_i$ of a certain Stixel $\stixeld*$, it becomes apparent that all other optimization parameters are independent of the actual choice of the plane parameters. We can thus simplify

\begin{align}
\argminx*{\planea*_i,\planeb*_i} \energyf{\stixel*{},\meas*{}} =
         \argminx*{\planea*_i,\planeb*_i} \energyf[stixel]{\stixeld*,\meas*{}} + \edepthmodel{\stixeld*}
        \punctspace .
\end{align}

Thus, we minimize the global energy function with respect to the plane parameters of all Stixels and all geometric classes independently. We can find an optimal solution of the resulting weighted least squares problem in closed form. However, we still need to compare the Stixel measurements to our new plane depth model. Therefore, the complexity added to the original formulation is another quadratic term in the image height.

\subsection{Stixel Cut Prior}
\label{subsec:cutprior}
The Stixel inference process described so far requires the estimation of the cost for each possible Stixel in an image. However, many Stixels can be trivially discarded, \eg in image regions with homogeneous depth and semantic input, making it possible to avoid the computation steps associated to the calculation of these. 

We propose a novel prior that exploits hypothesis generation to significantly reduce the computational burden of the inference task. To this end, we formulate a new prior similar to \cite{Cordts2014}; however, instead of Stixel bottom and top probabilities, we incorporate generic likelihoods for pixels being the cut between two Stixels.

We leverage this additional information adding a novel prior term for a Stixel $\stixeld*$
\begin{equation}
\energyf[cut]{\stixeld*} =
        \func{-\log}{\func{\cutpriorval*{\rowd*}}{cut}}
\end{equation}
where $\func{\cutpriorval*{\rowd*}}{cut}$ is the confidence for a cut at $\rowd*$, thus $\func{\cutpriorval*{\rowd*}}{cut} = 0$ implies that there is no cut between two Stixels at row $\row$.

As described in \cite{PfeifferDiss}, we can design a recursive definition of the optimization problem in order to solve the problem using a Dynamic Programming scheme. In order to simplify our description, we use a special notation to refer to Stixels: $ob_{b}^{t} = \{v^{b}, v^{t}, object\}$. Similarly, $OB^k$ is defined as the minimum aggregated cost of the best segmentation from position $0$ to $k$. The Stixel at the end of the segmentation associated with each minimum cost is denoted as $ob^k$. We next show a recursive definition of the problem:
\begin{equation}
\begin{split}
OB^k = min \begin{cases} E_{data}(ob_0^k)&+E_{prior}(ob_0^k)\\
E_{data}(ob_x^k)&+E_{prior}(ob_x^k,ob^{x-1}) \\ & + OB^{x-1}  \forall x \in cuts, x \leq k \\
E_{data}(ob_x^k)&+E_{prior}(ob_x^k,gr^{x-1}) \\ & + GR^{x-1}  \forall x \in cuts, x \leq k \\
E_{data}(ob_x^k)&+E_{prior}(ob_x^k,sk^{x-1}) \\ & + SK^{x-1}  \forall x \in cuts, x \leq k
\end{cases}
 \punctspace .
\end{split}
\label{eq:dyn_programming}
\end{equation}
We only show the case for \textit{object} Stixels, but the other cases are solved similarly. Also, $GR^k$ and $SK^k$ stand for \textit{ground} and \textit{sky} respectively. The base case problem, \ie segmenting a column of the single pixel at the bottom, is defined: $OB^0 = E_{data}(ob_0^0)+E_{prior}(ob_0^0)$. Our method trusts that all the optimal cuts will be included in our over-segmentation ($cuts$ in \cref{eq:dyn_programming}), therefore, only those positions are checked as Stixel bottom and top. This reduces the complexity of the Stixel estimation problem for a single column to $\mathcal{O}(h' \times h')$, where $h'$ is the number of over-segmentation cuts computed for this column, $h$ is image height and $h' \ll h$.

The computational complexity reduction becomes apparent in \cref{fig:stixel_graph_fig}.
As stated in \cite{Cordts2017}, the inference problem can be interpreted as finding the shortest path in a directed acyclic graph. Our approach prunes all the vertices associated with the Stixel's top row not included according to the Stixel cut prior, \cf \cref{fig:pruned_stixel_graph}.


\newlength{\spNodeDist} 

\setlength{\spNodeDist}{2.5mm}

\begin{figure}[t]
\centering
\subfloat[Stixel graph representation]
{
\centering
\begin{tikzpicture}


\tikzset{math mode/.style = { %
        execute at begin node=$, %
        execute at end node=$ %
    }}
\tikzset{math mode small/.style = { %
        execute at begin node=$\scriptstyle, %
        execute at end node=$ %
        }}

\tikzstyle{state}  = [ circle,
                       draw=black,
                       inner sep=0.2pt,
                       minimum size=15pt,
                       math mode,
                       node distance=1.2\spNodeDist and 2\spNodeDist,
                       text height=1.ex,
                       text depth=.25ex,
                       fill=gray!25,
                       pattern color=gray!25,
                       line width=1.2pt
                     ]

\definecolor{supportC}{RGB}{128, 64,128}
\definecolor{verticalC}{RGB}{220, 20, 60}
\definecolor{skyC}{RGB}{ 70,130,180}

\tikzstyle{supportN}  = [ state ,
                          fill=supportC!50
                        ]
\tikzstyle{verticalN} = [ state ,
                          fill=verticalC!50
                        ]
\tikzstyle{skyN}      = [ state ,
                          fill=skyC!50
                        ]

\tikzstyle{dummy}     = [ state ,
                          fill=none,
                          draw=none
                        ]

\tikzstyle{info}      = [ dummy,
                          node distance=-0.1\spNodeDist,
                        ]

\tikzstyle{info2}      = [ dummy,
                          node distance=-0.8\spNodeDist,
                        ]

\tikzstyle{trans}     = [ ->,
                          line width=1.2pt
                        ]



\node[state                     ]  (source)  {};

\node[verticalN, right=of source]  (v0)      {\object};
\node[supportN , above=of v0    ]  (s0)      {\ground};
\node[skyN     , below=of v0    ]  (k0)      {\sky};

\node[verticalN, right=of v0    ]  (v1)      {};
\node[supportN , above=of v1    ]  (s1)      {};
\node[skyN     , below=of v1    ]  (k1)      {};

\node[verticalN, right=of v1    ]  (vh)      {};
\node[supportN , above=of vh    ]  (sh)      {};
\node[skyN     , below=of vh    ]  (kh)      {};

\node[state    , right=of vh    ]  (sink)    {};

\node[info     , below=of k0    ]  (row1)    {1};
\node[info     , below=of k1    ]  (row2)    {2};
\node[info     , below=of kh    ]  (rowh)    {3};

\node[info2    , below=of row1  ]  (cut1)    {1};
\node[info2    , below=of rowh  ]  (cuth)    {1};
\node[info2    , below=of row2  ]  (cut2)    {1};

\node[info     , left =of row1  ]  (row)     {\rowtopd*=};
\node[info2    , left =of cut1  ]  (cut)     {\func{\cutpriorval*{\rowd*}}{cut}=};


\draw[trans] (source) -- (v0);
\draw[trans] (source) -- (s0);
\draw[trans] (source) -- (k0);
\draw[trans] (source) -- (s1);
\draw[trans] (source) .. controls ($(s0) + (-0,  3.5\spNodeDist)$) .. (sh);

\draw[trans] (v0)     -- (s1);
\draw[trans] (s0)     -- (s1);
\draw[trans] (k0)     -- (s1);

\draw[trans] (v0)     -- (sh);
\draw[trans] (s0)     .. controls ($(s1) + (0, 1.5\spNodeDist)$) .. (sh);
\draw[trans] (k0)     .. controls ($(v1) + (0.2, -1.6\spNodeDist)$) .. (sh);

\draw[trans] (v1)     -- (sh);
\draw[trans] (s1)     -- (sh);
\draw[trans] (k1)     .. controls ($(v1) + (0.5, -1.6\spNodeDist)$) .. (sh);

\draw[trans] (vh)     -- (sink);
\draw[trans] (sh)     -- (sink);
\draw[trans] (kh)     -- (sink);

\end{tikzpicture}
\label{fig:stixel_graph}
}

\subfloat[Pruned graph using Stixel cut prior]
{
\centering
\begin{tikzpicture}


\tikzset{math mode/.style = { %
        execute at begin node=$, %
        execute at end node=$ %
    }}
\tikzset{math mode small/.style = { %
        execute at begin node=$\scriptstyle, %
        execute at end node=$ %
        }}

\tikzstyle{state}  = [ circle,
                       draw=black,
                       inner sep=0.2pt,
                       minimum size=15pt,
                       math mode,
                       node distance=1.2\spNodeDist and 2\spNodeDist,
                       text height=1.25ex,
                       text depth=.25ex,
                       fill=gray!25,
                       pattern color=gray!25,
                       line width=1.2pt
                     ]

\definecolor{supportC}{RGB}{128, 64,128}
\definecolor{verticalC}{RGB}{220, 20, 60}
\definecolor{skyC}{RGB}{ 70,130,180}

\tikzstyle{supportN}  = [ state ,
                          fill=supportC!50
                        ]
\tikzstyle{verticalN} = [ state ,
                          fill=verticalC!50
                        ]
\tikzstyle{skyN}      = [ state ,
                          fill=skyC!50
                        ]

\tikzstyle{dummy}     = [ state ,
                          fill=none,
                          draw=none
                        ]

\tikzstyle{info}      = [ dummy,
                          node distance=-0.1\spNodeDist,
                        ]

\tikzstyle{info2}      = [ dummy,
                          node distance=-0.8\spNodeDist,
                        ]

\tikzstyle{trans}     = [ ->,
                          line width=1.2pt
                        ]



\node[state                     ]  (source)  {};

\node[verticalN, right=of source]  (v0)      {\object};
\node[supportN , above=of v0    ]  (s0)      {\ground};
\node[skyN     , below=of v0    ]  (k0)      {\sky};

\node[dummy    , right=of v0    ]  (dummyV)  {};
\node[dummy    , right=of s0    ]  (dummyS)  {};
\node[dummy    , right=of k0    ]  (dummyK)  {};

\node[verticalN, right=of dummyV]  (vh)      {};
\node[supportN , above=of vh    ]  (sh)      {};
\node[skyN     , below=of vh    ]  (kh)      {};

\node[state    , right=of vh    ]  (sink)    {};

\node[info     , below=of k0     ]  (row1)    {1};
\node[info     , below=of kh     ]  (rowh)    {3};
\node[info     , below=of dummyK ]  (dot2)    {2};

\node[info2     , below=of row1  ]  (cut1)    {1};
\node[info2     , below=of rowh  ]  (cuth)    {1};
\node[info2     , below=of dot2  ]  (cut2)    {0};

\node[info     , left =of row1  ]  (row)     {\rowtopd*=};
\node[info2    , left =of cut1  ]  (cut)     {\func{\cutpriorval*{\rowd*}}{cut}=};

\draw[trans] (source) -- (v0);
\draw[trans] (source) -- (s0);
\draw[trans] (source) -- (k0);
\draw[trans] (source) .. controls ($(s0) + (-0,  3.5\spNodeDist)$) .. (sh);
\draw[trans] (v0)     -- (sh);
\draw[trans] (s0)     -- (sh);
\draw[trans] (k0)     -- (sh);

\draw[trans] (vh)     -- (sink);
\draw[trans] (sh)     -- (sink);
\draw[trans] (kh)     -- (sink);

\end{tikzpicture}
\label{fig:pruned_stixel_graph}
}
\caption{Stixel inference illustrated as shortest path problem on a directed acyclic graph: the Stixel segmentation is computed by finding the shortest path from the source (left gray node) to the sink (right gray node).
The vertices represent Stixels with colors encoding their geometric class, \ie \textbf{g}round, \textbf{o}bject and \textbf{s}ky.
Only the incoming edges of ground nodes are shown for simplicity. Adapted from~\cite{Cordts2017}.
}
\label{fig:stixel_graph_fig}
\end{figure}
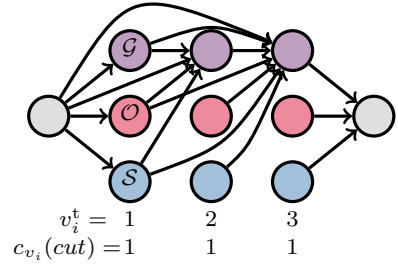
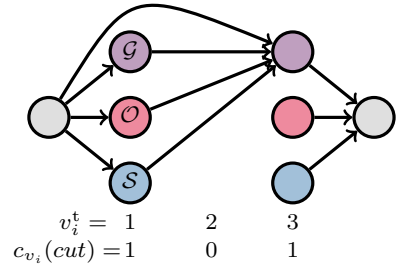


\section{Generation of the Stixel cut prior}
\label{sec:generate}

The previous section explained how to use a Stixel cut prior to reduce the computational complexity of the Stixel inference.
The idea is that many Stixel cuts could be trivially discarded, \eg in image regions with homogeneous depth and semantic input. We can save a lot of computation by not processing those unlikely Stixel cuts.
The goal is to devise a fast method to generate an over-segmentation of the optimal Stixel cuts. And, if those optimal cuts are included in the generated hypothesis, then the Stixel algorithm will provide the same output as in the original case, but doing much fewer computation steps.

We propose two methods to generate Stixel cuts.
The first method is a simple strategy that uses some mathematical concepts to identify points of interest \cf \cref{subsec:time_series}. It is a very fast approach, but misses some of the optimal Stixel cuts and, therefore, the final accuracy of the Stixel inference is reduced.
The second method uses a shallow Fully Convolutional Network (FCN) that is trained on the disparity map to infer likely Stixel cuts \cf \cref{subsec:cnn_presegmentation}. This strategy is also very fast, since the FCN is small, and is able to provide almost all of the optimal Stixel cuts.
For both methods, we leverage semantic segmentation information by including the edges of the semantic image into the set of the generated Stixel cuts.

\subsection{Time Series Compression}
\label{subsec:time_series}

The first method to generate Stixel cuts is based on the work of \cite{Oana2016}, and has linear time complexity and linear memory requirements.
In their work, each column of the disparity map is treated independently as a time series, \ie a signal with measurements on equal intervals of time.
They first perform an \textit{extreme points detection} step that generates a list of possible Stixel cuts, and then apply subsequent filters to this list in order to generate the final Stixel segmentation.
As we want to obtain an over-segmentation containing all the optimal Stixel cuts, we only use the first step of their proposal.

The detection of extreme points is based on techniques for time series compression \citep{fink2011compression}.
A time series can be compressed by selecting local extreme points, \ie maxima and minima of a function within a range. The assumption is that local extreme points are enough to find the important parts of the signal, and the rest would be unimportant points or noise.

In \cite{Oana2016} only left and right extrema are selected, while other kinds of extrema are discarded. Given a time series $\{t_1,t_2,\dotsc,t_i,\dotsc,t_{n-1},t_n\}$ and point $t_i$ with $1 < i < n$, the definition of left and right minimum is as follows (the definition of maxima is symmetric):

\begin{itemize}
\item $t_i$ is left minimum if $t_i < t_{i-1}$ and there is $t_j$ such that $j > i$ and $t_i = \dotsc = t_j < t_{j+1}$.
\item $t_i$ is right minimum if $t_i < t_{i+1}$ and there is $t_j$ such that $j < i$ and $t_{j-1} > t_j = \dotsc = t_i$.
\end{itemize}

\begin{figure}[ht]
\centering
\includegraphics[width=1\linewidth]{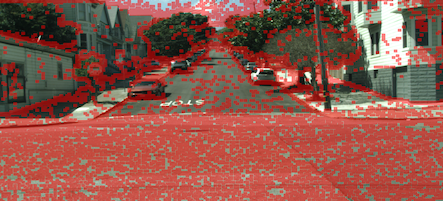}
\caption{Generated Stixel cuts (highlighted in red) using the left and right extrema as defined in \cite{Oana2016}, and also cuts generated from semantic segmentation. Stixel cut density is $30\%$, equivalent to a $3.3\times$ reduction in vertical resolution.}
\label{fig:timesseries_output}
\end{figure}

Similarly, we generate Stixel cuts by finding left and right extrema and the first and last points of the sequence of pixels in the column.
The example in \cref{fig:timesseries_output} illustrates the method.
The predicted Stixel cuts are indicated in red color.
In the example the vertical resolution is reduced around 3.3 times, which implies reduced computational work for the Stixel inference task.

\subsection{FCN-based method}
\label{subsec:cnn_presegmentation}

\begin{figure}[ht]
\centering
\includegraphics[width=1\linewidth]{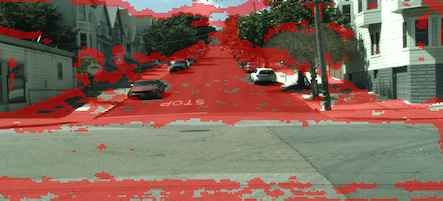}
\caption{Generated Stixel cuts (highlighted in red) for the FCN-based method. Stixel cut density is $31.5\%$, equivalent to a $3.2\times$ reduction in vertical resolution.}
\label{fig:fcn_output}
\end{figure}

We propose a novel shallow deep neural network \cf \cref{fig:fcn} that generates a set of promising Stixel cuts from depth images \cf \cref{fig:fcn_output}. We follow the proposal in \cite{JaschWR18}: we use disparities instead of depth.
We have experimentally found that adding the RGB image to the input of the neural network does not improve the accuracy of the method, compared to the simpler and faster strategy of directly adding the edges of the semantic image into the set of the generated Stixel cuts.

We design the network to provide an over-segmentation of the optimal Stixel cuts that should be significantly smaller than the total number of potential Stixel cuts (which is the height of the image).
Also, the computational work required for the network inference must be small, ideally similar to the Time Series method proposed in \cref{subsec:time_series}.
In the remainder of this section, we will first discuss the proposed network architecture, and then describe the data and training strategy.

\subsubsection{Network architecture}

\begin{figure}
\centering
\includegraphics[width=\textwidth,height=0.7\textheight,keepaspectratio]{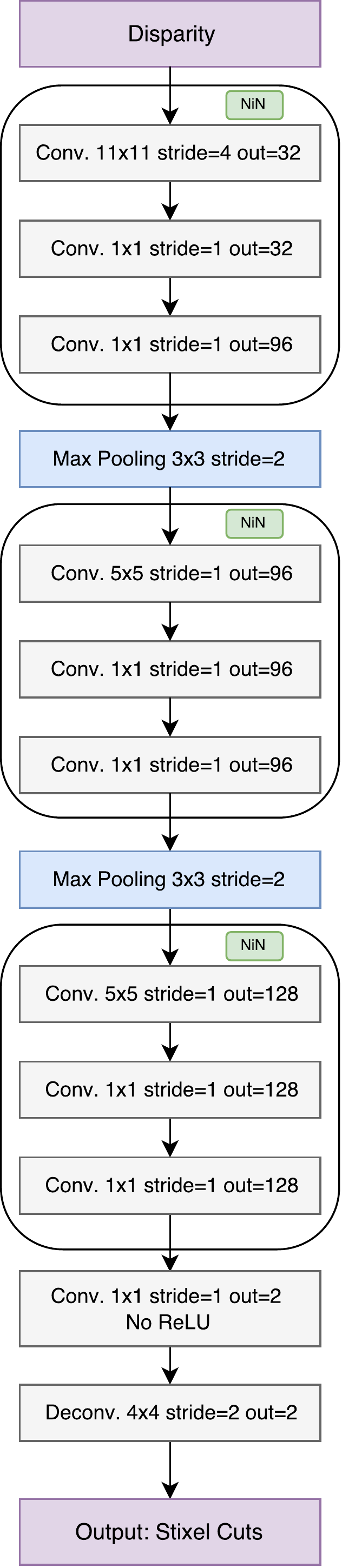}
\caption{Definition of the proposed Fully Convolutional Network for generating Stixel cuts.}
\label{fig:fcn}
\end{figure}

Our proposal is based on the architecture described by \cite{schneider2017multimodal}.
They present a multi-modal FCN designed for semantic segmentation with a mid-level fusion architecture that exploits complementary input cues, \ie RGB and disparity data.
Their design includes the Network in Network (NiN) method proposed by \cite{lin2013}.
Our proposal inherits the network branch that processes the disparity data and discards the branch on the RGB data, which is described in detail in \cref{fig:fcn}.
The proposed FCN is a very shallow network with three consecutive NiNs, and a final deconvolution that recovers the desired resolution of the Stixel cuts.
The output of the FCN is a binary image indicating whether or not there is a Stixel cut for that pixel.


\subsubsection{Training data}

\begin{figure}
\centering
\includegraphics[width=1\linewidth]{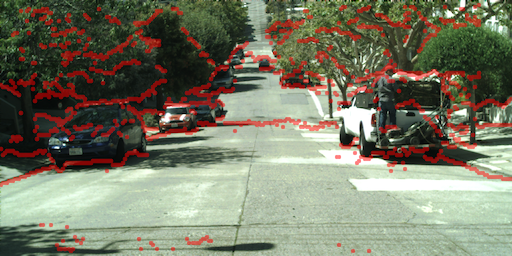}
\caption{Sample image from the real-data sequence used for Stixel cut generation. Stixel cut ground-truth is highlighted in red.}
\label{fig:sf_dataset}
\end{figure}

We trained the proposed FCN using disparity maps generated from images in the Synthia synthetic dataset \citep{RosCVPR16} and from images in a real-data sequence (6757 images) recorded in San Francisco, \cf \cref{fig:sf_dataset}. In both cases, the disparity maps are generated from the left and right RGB images using a stereo matching algorithm
\citep{Hirschmuller2008}. This is the expected situation in a realistic scenario, where the SGM algorithm in the perception pipeline generates the disparity map and feeds the FCN that produces the Stixel over-segmentation.

The ground-truth for the training data (the expected Stixel cuts) is generated as a combination of methods. In the case of the annotated synthetic dataset, which contains both pixel-level semantic and instance-level annotations, the ground-truth includes, as desired Stixel cuts, the boundaries of the instances and the semantic classes in the image (as in \cite{Cordts2017}).
Finally, the Stixel cuts associated to disparity changes are obtained by running the Stixel inference method. In the real-data sequence, we only perform this last step because we lack ground-truth.

As discussed previously \cf \cref{subsubsec:model_complexity_prior}, the definition of the parameters of the Stixel model represent a trade-off between compactness and accuracy. Since we need an over-segmentation of the optimal Stixel cuts, we adjust the parameters of the model to be conservative and to favor accuracy versus compactness.

The idea of using the Stixel model as a way to train a fast and simple neural network to approximate the optimal Stixel segmentation is inspired by model distillation techniques \citep{BucilaCN06}.
The comparatively slow Dynamic Programming method to solve the probabilistic model is used to transfer the knowledge inside the complex model to a fast and compact FCN that approximates the optimal Stixel cuts.

\subsubsection{Training strategies}

Since our problem is to classify each pixel of our input disparity map as \textit{cut} or \textit{not-cut}, we use cross-entropy as the loss function that must be minimized.
The distribution of \textit{cut}/\textit{not-cut} is strongly biased in our input and, accordingly, we introduce a class-balancing weight in the loss function, similarly to \cite{XieT17}.
These weights cause the FCN to generate wider edges \cf \cref{fig:fcn_output}.
This is useful, since the FCN roughly detects the Stixel cut positions, and the precise detection is left to the Stixel inference.

We set the learning rate to $10^{-8}$ and the batch size to five: four of those inputs are Synthia images and one of them is a real-data image.
The missing disparities are encoded as $-1$.
Input normalization is done by subtracting the mean value from the disparity map.
We initialize the FCN with the weights used in \cite{schneider2017multimodal}, since semantic segmentation is a similar problem.

\section{Experiments}
\label{sec:experiments}
This section assesses the accuracy and run-time of our proposal. A previous concern is to verify that our method not only improves the representation of scenes with non-flat roads, but also maintains the accuracy for scenes containing only flat roads. For that purpose, we present datasets of synthetic and real data to evaluate our proposal in \cref{subsec:datasets}. 
We introduce inputs, metrics, baselines, and other experimental details in \cref{subsec:experiment_details}. Finally, quantitative and qualitative results are reported in \cref{subsec:results}.

\subsection{Datasets}
\label{subsec:datasets}

\begin{figure*}[ht]
  \centering
  \subfloat
  {
    \centering
    \includegraphics[trim=25 50 64 25,clip,width=0.32\textwidth,height=2.8cm]{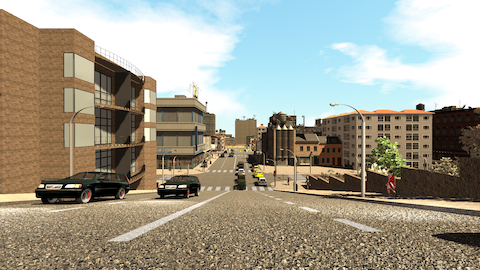}
  }
  \subfloat
  {
    \centering
    \includegraphics[trim=100 200 250 100,clip,width=0.32\textwidth,height=2.8cm]{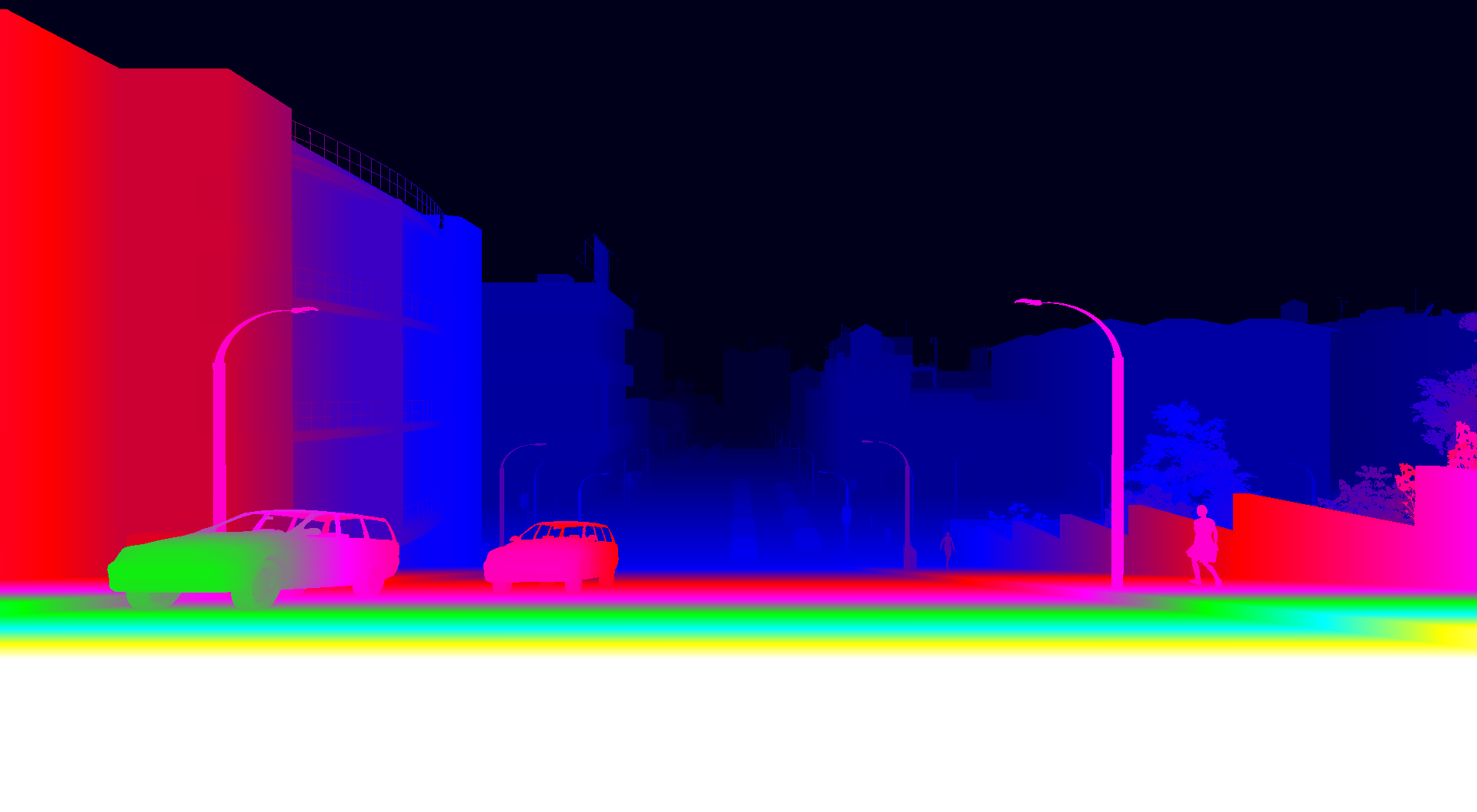}
  }
  \subfloat
  {
    \centering
    \includegraphics[trim=100 200 250 100,clip,width=0.32\textwidth,height=2.8cm]{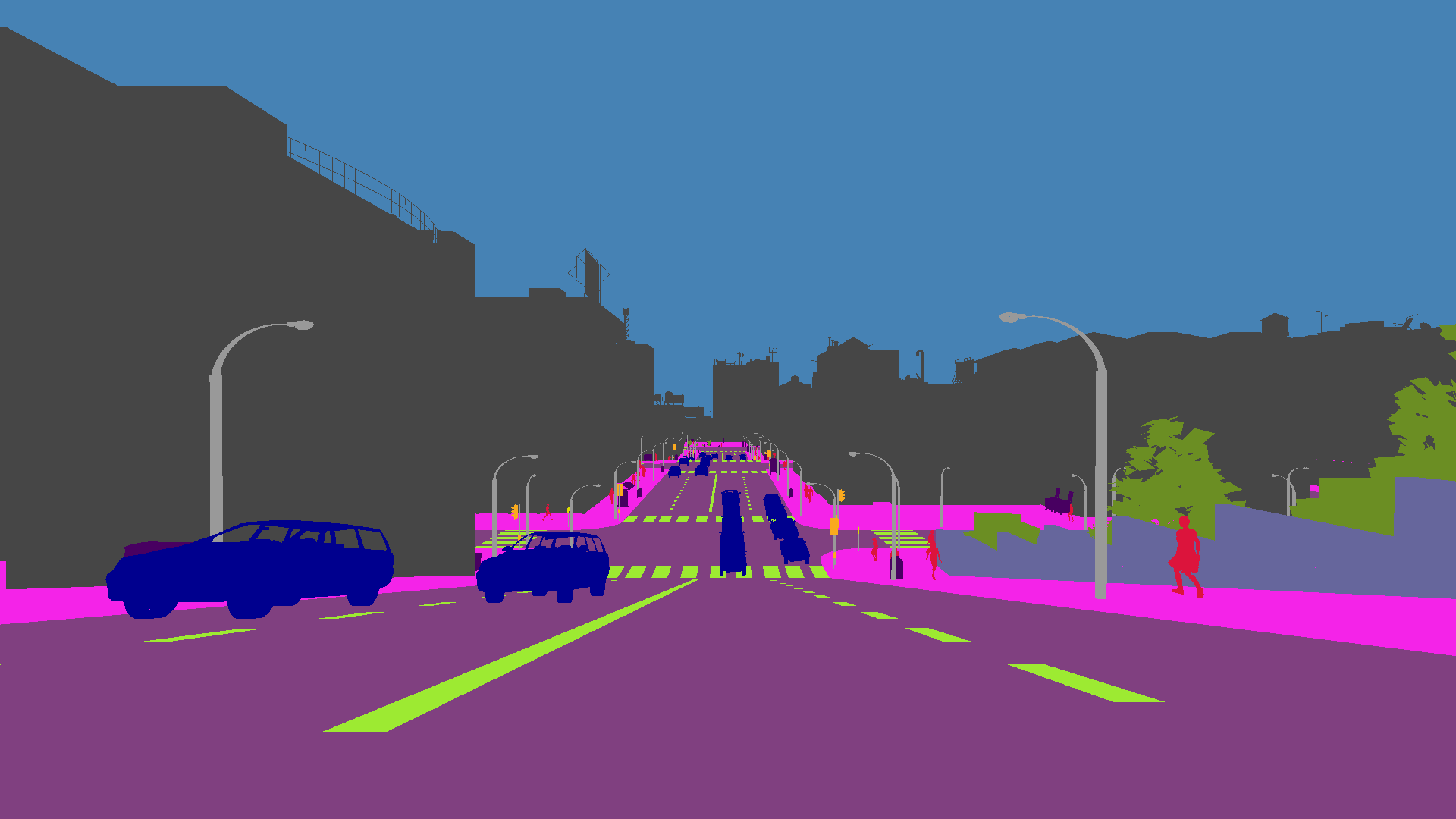}
  }
  
  \subfloat
  {
  \begin{tabularx}{1\textwidth}{XXXXXXXX}
    \labelcolorb{sidewalk}      &
    \labelcolor{building} &
    \labelcolor{vegetation}    &
    \labelcolorb{traffic light} &
    \labelcolorb{traffic sign}  &
    \labelcolor{bicycle} &
    \labelcolor{motorcycle} &
    \labelcolorb{road lines}
  \end{tabularx}
  }\\[0.01pt]
  
  \subfloat
  {
  \begin{tabularx}{1\textwidth}{XXXXXXXXXXXX}
    \labelcolorb{terrain} &
    \labelcolor{road}          &
    \labelcolor{wall} &
    \labelcolor{pole} &
    \labelcolorb{rider} &
    \labelcolor{truck} &
    \labelcolor{bus} &
    \labelcolor{train} &
    \labelcolorb{fence} &
    \labelcolor{person}        &
    \labelcolor{sky} &
    \labelcolor{car}
  \end{tabularx}
  }
\caption{The \datasetname{} Dataset. A sample frame (left) with its depth (center) and semantic labels (right).}
\label{fig:synthia-sf}
\end{figure*}

As our Stixel model represents geometric and semantic information, we must evaluate the accuracy of our method for both. For that purpose, we select \textit{Ladicky} \citep{Ladicky2014}, an annotated subset of \textit{KITTI} \citep{Geiger2012CVPR}, which is, to the best of our knowledge, the only dataset containing both dense semantic labels and depth ground-truth. \revs{It consists of} a set of 60 images with 0.5 MP resolution that we use for evaluating Stixel semantic and depth accuracy. We follow the suggestion given by the author \citep{Ladicky2014} to ignore the three rarest object classes, which leaves us with 8 classes. 

\revs{Additionally, for training our semantic segmentation FCN, we use publicly available semantic annotations on other parts of \textit{KITTY} \citep{KunduLDLR14,HeU13,SenguptaGST13,XuDBZD13,ZhangCVZ15}}. Our total training set is composed of 676 images, where we harmonized the object classes used by the different authors \revs{to the previously mentioned set suggested by \cite{Ladicky2014}. This harmonization and data processing is the same applied in the previous work \citep{Schneider2016} to allow for fair comparison.}


In order to further evaluate disparity accuracy we use the training data of the well-known stereo challenge \textit{KITTI 2015} \citep{Geiger2012CVPR}. This dataset provides a set of 200 images with sparse disparity ground-truth obtained from a laser scanner. There is no suitable semantic ground-truth available for this dataset.

Furthermore, we also evaluate semantic accuracy using \textit{Cityscapes} \citep{Cordts2016}, a highly complex dataset with dense annotations of 19 classes on $\sim3000$ images for training and 500 images for validation that we use for testing.

Unfortunately, all the above datasets were generated in flat road environments. Hence, they only help us validate that we are not decreasing our accuracy for this kind of environments. In order to compare the accuracy of competing algorithms on non-flat road scenarios, we need a new dataset.

Therefore, we introduce a new synthetic dataset inspired by \cite{RosCVPR16}. This dataset has been generated with the purpose of evaluating our proposed model; however, it contains enough information to be useful in additional related tasks, such as object recognition, semantic and instance segmentation, among others.

\datasetnamefull{} (\textit{\datasetname}) consists of photo-realistic frames rendered from a virtual city and comes with precise pixel-level depth and semantic annotations for 19 classes \cf \cref{fig:synthia-sf}. This new dataset contains 2224 images that we use to evaluate both depth and semantic accuracy in non-flat roads.

\begin{figure}[ht]
\begin{tikzpicture}
  \begin{axis}[
    height=5cm,
    width=1\columnwidth,
    ybar=5pt,
    enlargelimits=0.15,
    legend style={at={(0.5,-0.15)},
       anchor=north,legend columns=-1},
    ylabel={Frame-rate (Hz)},
    symbolic x coords = {No over-segmentation, Time Series, FCN},
    xtick=data, 
    nodes near coords, 
    nodes near coords align={vertical}, 
  ]
  \addplot coordinates { (No over-segmentation,19.4) (Time Series,38.9) (FCN,54.3) };
  \addplot coordinates { (No over-segmentation,4.7) (Time Series,33.1) (FCN,35.2)};
  \legend{Semantic Stixels, Ours}
  \end{axis}
\end{tikzpicture}
\caption{Frame-rate of our method \rev{(only the Stixel computation step and the corresponding over-segmentation approach)} compared to Semantic Stixels \citep{Schneider2016} for \datasetname{} (image resolution of 1920$\times$1080) on a multi-threaded CPU implementation (Intel i7-6800K) computed with a Stixel width of 8 pixels and equivalent down-sampling in the v-direction. Different methods of over-segmentation are also compared, these are: Time Series \cf \cref{subsec:time_series}, FCN \cf \cref{subsec:cnn_presegmentation}.}
\label{fig:frames_per_second}
\end{figure}
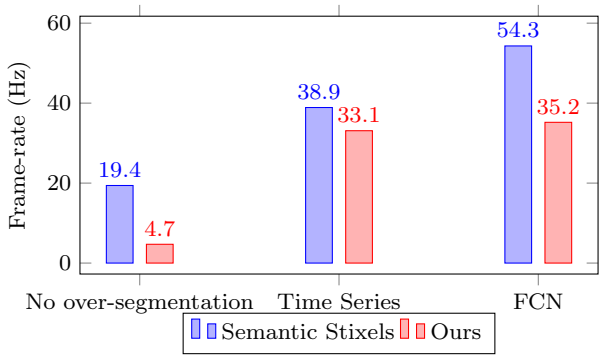

\begin{table}[ht]
\centering
\caption{\rev{Per-stage report of frame-rate of our pipeline for a stereo pair of resolution 1242 $\times$ 375. \textit{OS} stands for \textit{Over-segmentation}. \textit{SGM} run-time using a CPU Intel i7-6800K. For the \textit{Semantic Segmentation} method, a Maxwell NVidia Titan X is used. Note that Stixel frame-rate is variable if we use an over-segmentation method, therefore we provide a representative run-time. The total frame-rate is reported as the sum of the stages.}}

\begin{tabular}{l|c}
\toprule
Stage & Frame-rate (Hz) \\
\midrule
SGM    &  55     \\
\hline
Semantic Segmentation & 47.6 \\
\hline
Our Stixels (OS: Time Series) & 116 \\
\hline
Our Stixels (OS: FCN) & 130 \\
\hline
Our Stixels (No OS) & 61 \\
\bottomrule
Total (OS: Time Series) & 20.92 \\
\hline
Total (OS: FCN) & 21.33 \\
\hline
Total (No OS) & 18 \\
\bottomrule
\end{tabular}

\label{table:per_stage_timming}
\end{table}

\begin{table*}[ht]
\centering
\caption{Accuracy of our methods compared to Semantic Stixels \citep{Schneider2016}, raw SGM and FCN. We evaluate on four datasets: Ladicky \citep{Ladicky2014}, KITTI 15 \citep{Geiger2012CVPR}, Cityscapes \citep{Cordts2016} and \datasetname{} using these metrics: Disparity Error (less is better) and Intersection over Union (more is better) \cf \cref{subsec:datasets} and \cref{subsec:metrics}. \textit{Fast} versions are detailed in \cref{subsec:time_series} and \cref{subsec:cnn_presegmentation}. Significantly best results are highlighted in bold.}

\begin{tabular}{ll|cc|cc|cc|cc}
\toprule
 & & \multicolumn{2}{c|}{\textbf{Input}}  & \multicolumn{2}{c|}{\textbf{No over-segmentation}}  & \multicolumn{2}{c|}{\textbf{Fast: Time Series}} & \multicolumn{2}{c}{\textbf{Fast: FCN}}     \\
Metric                           & Dataset      & SGM    & FCN  & Sem. Stixels  & Ours  & Sem. Stixels & Ours & Sem. Stixels & Ours     \\
\midrule
\multirow{3}{*}{Disp Error (\%)} & Ladicky      & 16.66  & -    & 17.38 & 16.84 & 17.60       & 17.01       & 17.44           & 16.84               \\
                                 & KITTI 15     & 11.01  & -    & 11.05 & 11.21 & 11.9        & 11.9        & 11.21           & 11.24               \\
                                 & \datasetname & 11.06 & -    & 29.33 & \textbf{12.99} & 30.60       & 14.20       & 31.12           & 14.19              \\
\hline
\multirow{3}{*}{IoU (\%)}        & Ladicky      & -      & 69.8 & 66.2  & 66.1  & 66.0        & 66.0        & 66.2            & 66.1                \\
                                 & Cityscapes   & -      & 66.7 & 65.4  & 65.8  & 64.9        & 65.0        & 65.5            & 65.6                \\
                                 & \datasetname & -      & 48.1 & 46.0  & \textbf{48.5}  & 45.7        & 48.0        & 47.0            & \textbf{48.6}                \\
\bottomrule
\end{tabular}

\label{table:accuracy_results}
\end{table*}

\begin{table*}[ht]
\centering
\caption{Number of Stixels ($10^3$) generated by our methods compared to Semantic Stixels \citep{Schneider2016} and raw input (total number of pixels). We evaluate on four datasets: Ladicky \citep{Ladicky2014}, KITTI 15 \citep{Geiger2012CVPR}, Cityscapes \citep{Cordts2016} and \datasetname{} \cf \cref{subsec:datasets}. \textit{Fast} versions are detailed in \cref{subsec:time_series} and \cref{subsec:cnn_presegmentation}.}

\begin{tabular}{l|c|cc|cc|cc}
\toprule
 & Input & \multicolumn{2}{c|}{No over-segmentation}  & \multicolumn{2}{c|}{Fast: Time Series} & \multicolumn{2}{c}{Fast: FCN}     \\
Dataset      & SGM/FCN  & Sem. Stixels  & Ours  & Sem. Stixels & Ours & Sem. Stixels & Ours     \\
\midrule
Ladicky      & 454  & 0.6   & 0.6   & 0.6         & 0.6         & 0.6             & 0.6                 \\
KITTI 15     & 452  & 0.7   & 0.7   & 0.7         & 0.7         & 0.7             & 0.7                 \\
Cityscapes   & 2 k  & 1.4   & 1.5   & 1.3         & 1.4         & 1.4             & 1.5                 \\
\datasetname & 2 k  & 1.5   & 1.7   & 1.2         & 1.3         & 1.3             & 1.3                \\
\bottomrule
\end{tabular}
\label{table:nstixels_results}
\end{table*}

\subsection{Experiment details}
\label{subsec:experiment_details}

\subsubsection{Metrics}
\label{subsec:metrics}

We evaluate our proposed method in terms of semantic and depth accuracy using two metrics. The depth accuracy is obtained as the rate of outliers of the disparity estimates, the standard metric used to evaluate on KITTI benchmark \citep{Geiger2012CVPR}. An outlier is a disparity estimation with an absolute error larger than 3 px or a relative deviation larger than 5\% compared to the ground-truth. The semantic accuracy is evaluated with the average Intersection-over-Union (IoU) over all classes, which is also a standard measure for semantic segmentation \citep{Everingham2015}. We measure the number of Stixels generated per image to quantify the complexity of the obtained representation. Finally, we evaluate the inference speed of the algorithm using the Frame-rate (Hz) metric, which helps us estimate if our system is capable of real-time performance. 
\rev{All the execution times of \textit{Stixels} and \textit{SGM} are obtained using a multi-threaded implementation running on standard consumer hardware: Intel i7-6800K. The semantic segmentation FCN frame-rate estimations are obtained using Maxwell NVidia Titan X. The Stixel frame-rate includes the over-segmentation approach. Note that Stixel frame-rate is variable if we use an over-segmentation method, \ie it will depend on the number of Stixel cuts removed, therefore we provide a representative frame-rate. Similarly to \cite{Cordts2017}, we can maximize the throughput of the system by computing SGM and Semantic Segmentation in parallel, then the system would run with one frame delay.}

\subsubsection{Baseline}

Semantic Stixels \citep{Schneider2016} serve as our comparison baseline, as they achieve state-of-the-art results in terms of Stixel accuracy. We provide the accuracy of our new disparity model, \cf \cref{sec:method}. Finally, we evaluate the complexity of the fast approach defined in \cref{subsec:cutprior}, with the two over-segmentation techniques presented in \cref{subsec:time_series} and \cref{subsec:cnn_presegmentation}.

\subsubsection{Input}

As input, we use disparity images obtained via SGM \citep{Hirschmuller2008} and pixel-level semantic labels computed by an FCN \citep{Long2015}. We use the same FCN model used in \cite{Schneider2016} without retraining, to allow for comparison. For the same reason, we set Stixel width to 8 px. The same down-sampling is applied in the vertical direction. The rest of the parameters used are taken from \cite{Schneider2016}.

We use the camera parameters obtained after calibration to set the expected values of $\mu_{ground}^{\planea*}$ and $\mu_{ground}^{\planeb*}$. For \textit{object} Stixels, we set $\sigma_{object}^{\planeb*} \rightarrow 0, \mu_{object}^{\planeb*} = 0$ because the disparity is too noisy for the slanted object model. Finally, since \textit{sky} Stixels can not have slanted surfaces, we set: $\mu_{sky}^{\planea*}=0, \mu_{sky}^{\planea*}=0, \sigma_{sky}^{\planea*} \rightarrow 0, \sigma_{sky}^{\planeb*} \rightarrow 0$.

In order to improve the computational efficiency of our approach, we use the two \textit{Fast} Stixel over-segmentation methods presented in \cref{subsec:time_series}, labeled as \textit{Time Series}, and \cref{subsec:cnn_presegmentation}, labeled as \textit{FCN}.

\begin{figure*}[ht]
\centering
\captionsetup[subfigure]{position=top,labelformat=empty}
\subfloat[RGB Image]
{
  \centering
  \includegraphics[trim= 0 0 0 0.2cm, width=0.30\textwidth,height=3cm]{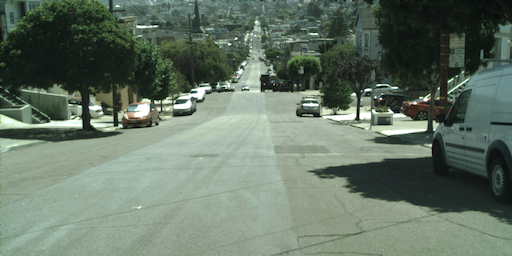}
}
\subfloat[Original Stixels \citep{Schneider2016}]
{
  \centering
  \includegraphics[width=0.30\textwidth,height=3cm]{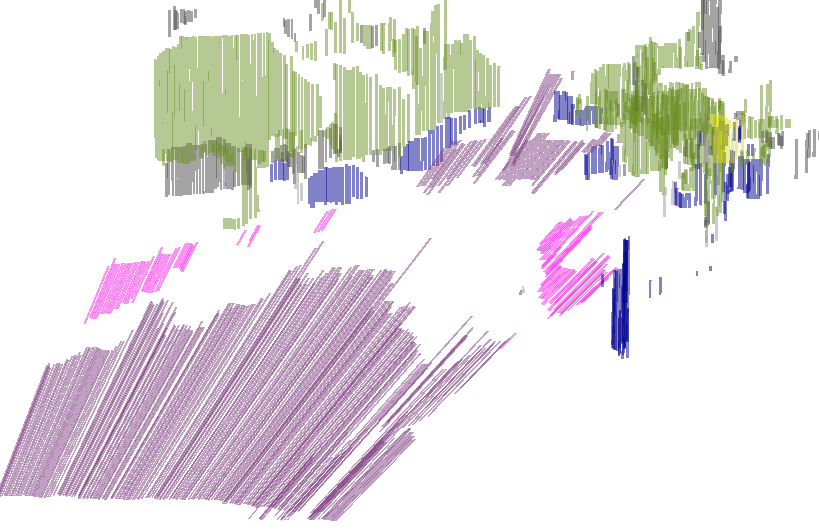}
}
\subfloat[Our Stixels]
{
  \centering
  \includegraphics[width=0.30\textwidth,height=3cm]{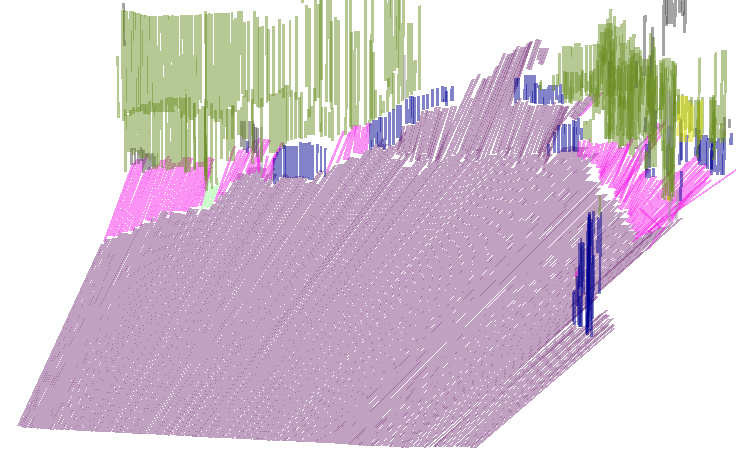}
}

\subfloat
{
  \centering
  \includegraphics[trim= 0 0 0 0.2cm, width=0.30\textwidth,height=3cm]{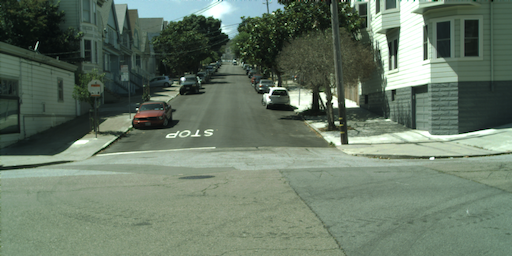}
}
\subfloat
{
  \centering
  \includegraphics[width=0.30\textwidth,height=3cm]{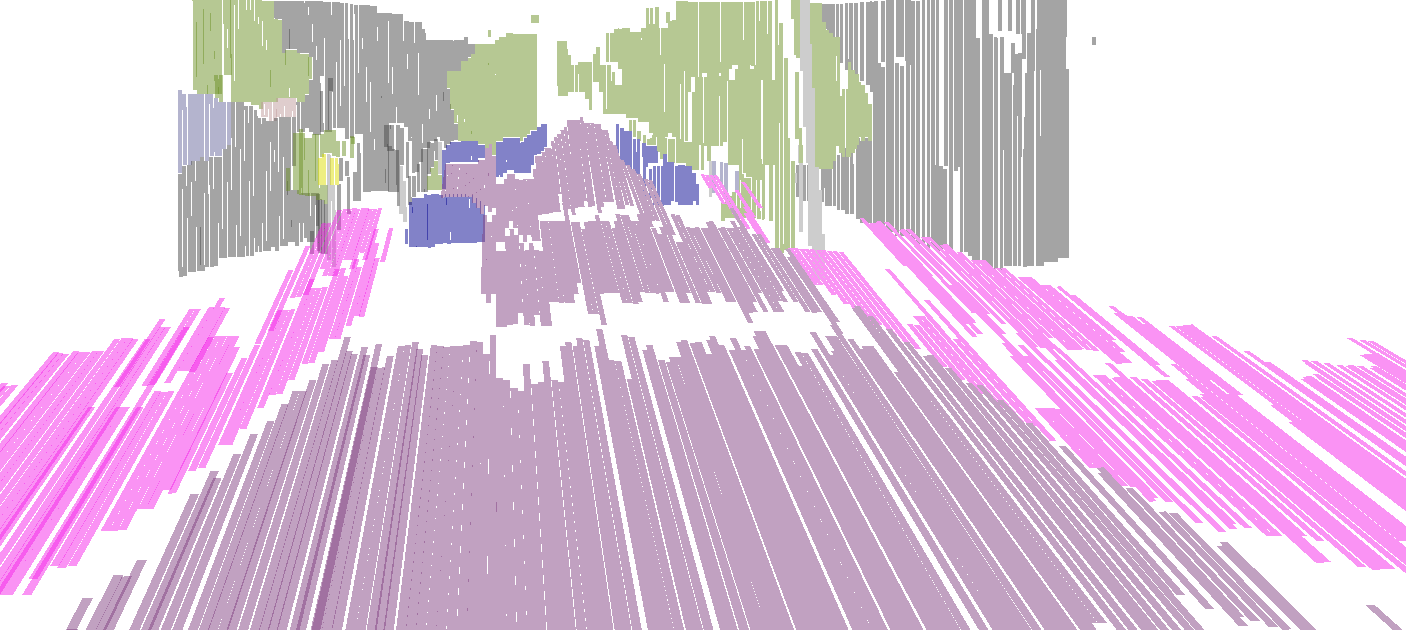}
}
\subfloat
{
  \centering
  \includegraphics[width=0.30\textwidth,height=3cm]{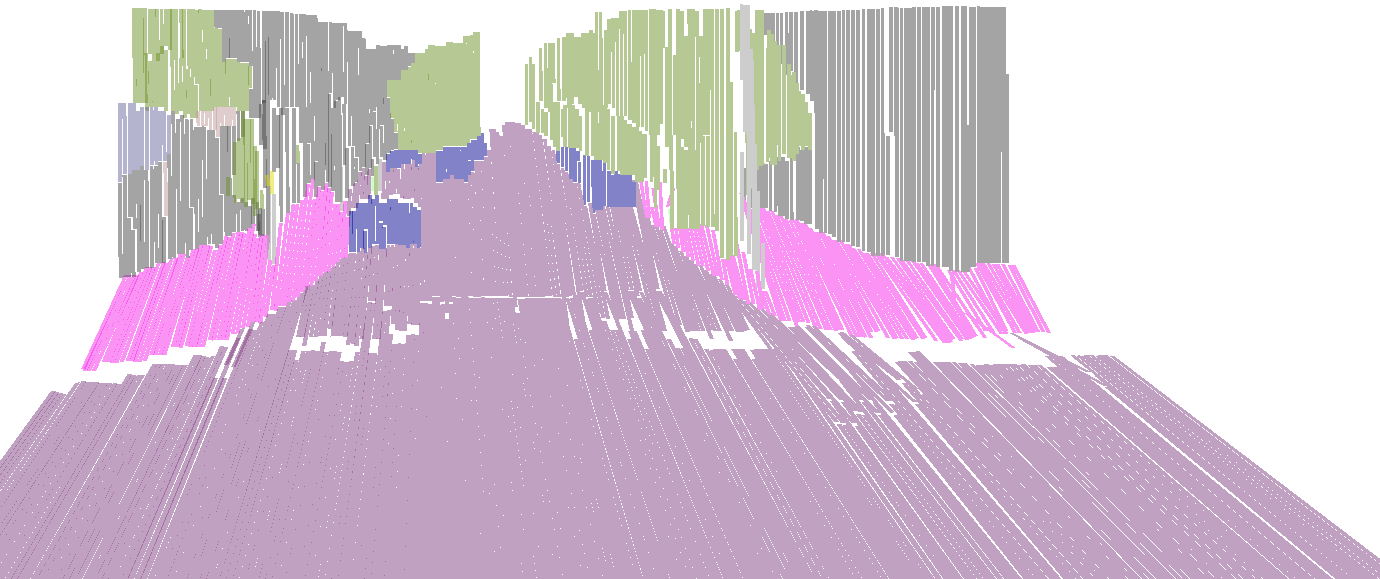}
}

\subfloat
{
  \centering
  \includegraphics[trim= 0 0 0 0.2cm, width=0.30\textwidth,height=3cm]{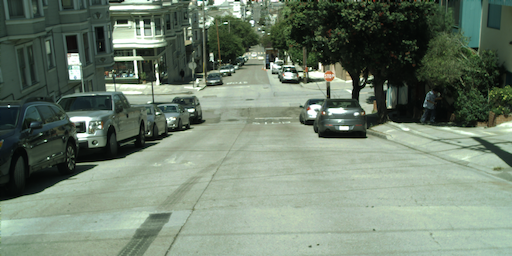}
}
\subfloat
{
  \centering
  \includegraphics[width=0.30\textwidth,height=3cm]{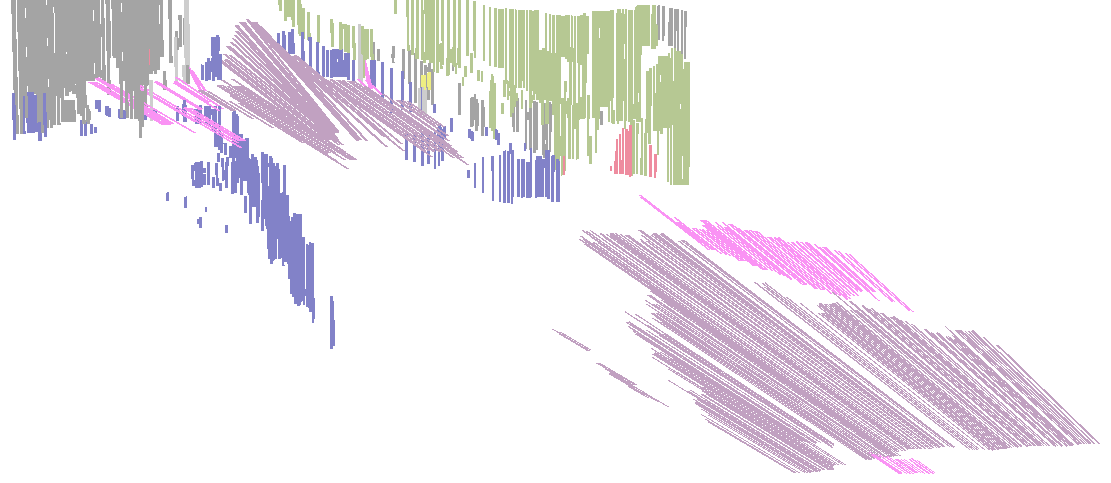}
}
\subfloat
{
  \centering
  \includegraphics[width=0.30\textwidth,height=3cm]{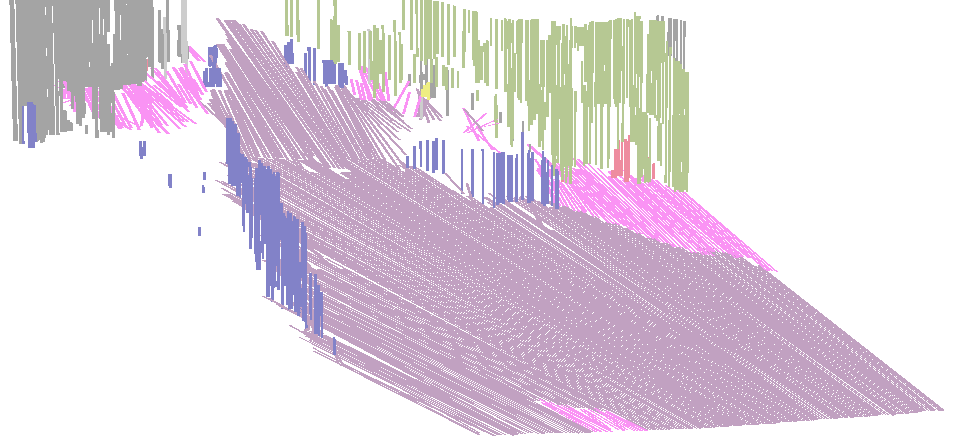}
}

\subfloat
{
  \centering
  \includegraphics[trim= 0 0 0 0.2cm, width=0.30\textwidth,height=3cm]{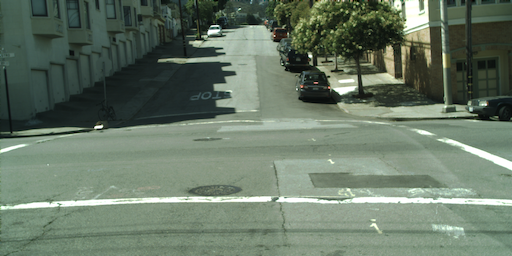}
}
\subfloat
{
  \centering
  \includegraphics[width=0.30\textwidth,height=3cm]{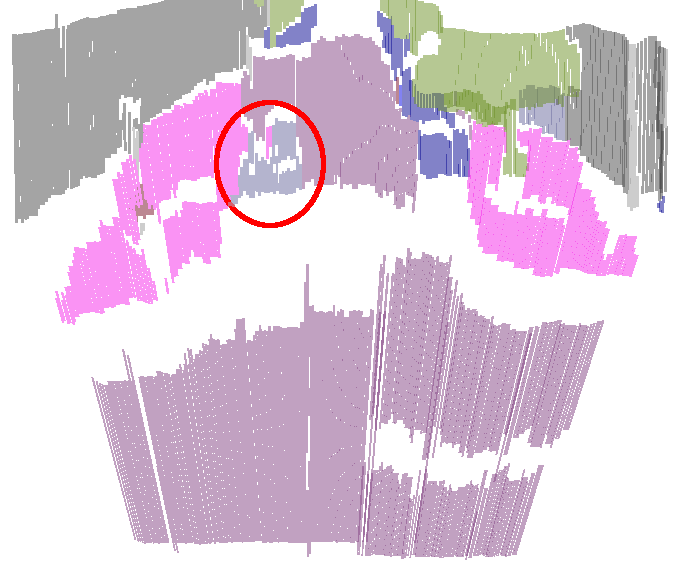}
}
\subfloat
{
  \centering
  \includegraphics[width=0.30\textwidth,height=3cm]{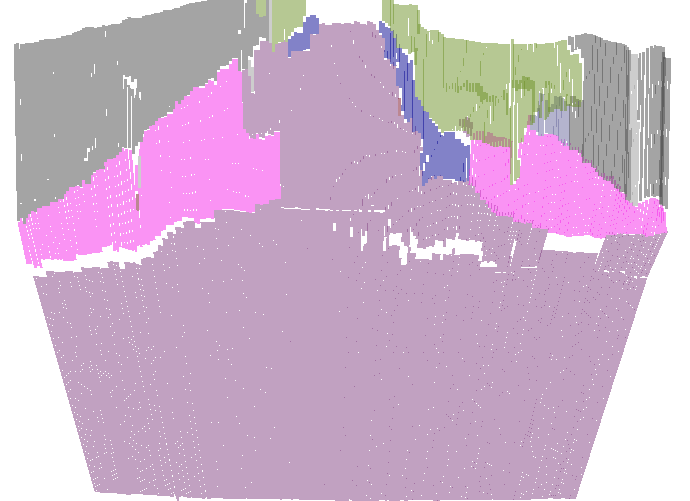}
}
\caption{Exemplary outputs on real data: in all cases with non-flat roads our model correctly represents the scene, while retaining accuracy on objects. The last example shows a failure case, where our approach classifies the road as sidewalk due to erroneous semantic input. However, the original approach reconstructs a wall in this case, highlighted by a red circle. This could lead to an emergency break.}
\label{fig:comparison}
\end{figure*}

\subsection{Results}
\label{subsec:results}

The quantitative results of our proposals and baselines, as described in \cref{sec:method}, are shown in \cref{table:accuracy_results,table:nstixels_results,fig:frames_per_second}.

The first observation is that our method achieves comparable or slightly better results on all datasets with flat roads \cf compare \textit{Semantic Stixels} to \textit{Ours} for \textit{Ladicky}, \textit{KITTI 15} and \textit{Cityscapes} datasets in \cref{table:accuracy_results}. These results indicate that the novel and more flexible model does not harm the accuracy in such scenarios.

We also observe that our novel model is able to accurately represent non-flat scenarios in contrast to the original Stixel approach, yielding a substantially increased depth accuracy of more than $16\%$ \cf when comparing \textit{Semantic Stixels} to \textit{Ours} for the \textit{\datasetname} dataset in \cref{table:accuracy_results}. \rev{Additionally, to verify that our method equally works also on real data, we provide a video of the Stixel 3D representation of a challenging non-flat road scene as supplementary material}. Results also improve in terms of semantic accuracy, which we explain as a consequence of the joint semantic and depth inference that benefits from a better depth model.

\rev{A perfect over-segmentation method would find all optimal cuts, and consequently, it would have the same accuracy as not using any over-segmentation.}

\rev{Our novel approach \textit{Fast: FCN} has an accuracy almost equal to not using any over-segmentation method (in all cases but one). Note that, our proposed approach \textit{Fast: FCN} is superior to \textit{Fast: Time Series} method in all cases \cf when comparing both methods for the \textit{\datasetname} dataset in \cref{table:accuracy_results}.}

\rev{Both over-segmentation methods increase the error for our challenging \textit{\datasetname} dataset; we think this is because of the difficult road Stixel cuts in these scenes, \cf compare \textit{No over-segmentation} to \textit{Fast} methods in \cref{table:accuracy_results}.}

All variants are compact representations of the surrounding, since the complexity of the Stixel representation is small compared to the high resolution input images, \cf \cref{table:nstixels_results}.

Our last observation is that the proposed \textit{Fast} variants improve the run-time of the original Stixel approach by up to $2\times$, and also improve the novel Slanted Stixel approach by up to $7\times$, with only a slight drop in depth accuracy \cf \cref{fig:frames_per_second}. The benefit increases with higher resolution input images due to the quadratic and cubic computational complexity of the original and slanted Stixel inference methods, respectively. \rev{We also detail per-stage run-time \cf \cref{table:per_stage_timming} for completeness.}


In addition to the quantitative evaluation presented before, we have visually inspected many of the obtained Stixel representations, to check the qualitative differences between our proposal and the previous work. 
\Cref{fig:comparison} illustrates some of these examples, in which the scenes with non-flat roads are correctly represented and all the objects in the scenario are identified by our proposal, while the previous model produces an incomplete road representation, or even generates non-existing objects at some road positions.

\section{Conclusions}
\label{sec:conclusions}

This paper presented a novel depth model for the Stixel world that is able to account for non-flat roads and slanted objects in a compact representation that overcomes the previous restrictive constant height and depth assumptions. This change in the way Stixels are represented is required for difficult environments that are found in many real-world scenarios. Moreover, in order to significantly reduce the computational complexity of the extended model, a novel approximation has been introduced that consists of checking only reasonable Stixel cuts inferred using fast methods. We showed in extensive experiments on several related datasets that our depth model is able to better represent slanted road scenes, and that our approximation is able to reduce the run-time drastically, with only a slight drop in accuracy.

\rev{As future work, we would like to focus on circumventing the limitations of our method. Namely, (1) the vertical/column independence assumed by the model is clearly not true. A more global representation, \eg super-pixels that span vertically and horizontally, would be more compact and less prone to errors; (2) some surfaces are not well represented by a linear model, \eg cars. A more complex depth model and specific models for each semantic class could represent more faithfully the scene. Nonetheless, a model with more free variables could also lead to a bad representation because of the noise; (3) the proposed over-segmentation algorithm has a non-predictable run-time. And this is a bad characteristic for a real-time system. The worst-case scenario, \ie no Stixel cuts removed, is as slow as not using over-segmentation at all (although very unlikely); (4) in case of movement of the stereo rig during operation, there could be an offset in roll effectively breaking the vertical world assumption.}

\begin{acknowledgements}
This work has been partially supported by Spanish TIN2017-84553-C2-1-R (MINECO/AEI/FEDER, UE). We also thank the Generalitat de Catalunya CERCA Program, the 2017-SGR-1597 and 2017-SGR-313 projects, as well as the ACCIO agency. We also acknowledge SEBAP for the internship funding program. Antonio M. L\'opez acknowledges the financial support by the Spanish TIN2017-88709-R (MINECO/AEI/FEDER, UE), and by ICREA under the ICREA Academia Program. Finally, we thank Francisco Molero and Marc Garc{\'i}a at CVC/UAB for the generation of the SYNTHIA-SF dataset.
\end{acknowledgements}

\bibliographystyle{spbasic}      
\bibliography{egbib}   

\end{document}